%% file: main.tex
\documentclass[10pt,journal,compsoc]{IEEEtran}

\usepackage{xcolor}
\usepackage{helvet}
\usepackage{courier}
\usepackage{graphicx}
\usepackage{amsmath,amsthm,amssymb}
\usepackage{textcomp,booktabs}
\usepackage{footmisc}
\usepackage[normalem]{ulem}
\usepackage{multirow}
\usepackage{setspace}
\usepackage{bigstrut}
\usepackage{array}
\usepackage{xcolor}
\usepackage{amsfonts}
\usepackage{bm}
\usepackage{url}
\usepackage{xcolor}
\usepackage{helvet}
\usepackage{courier}
\usepackage{subfigure}
\usepackage{balance}
\usepackage{amsmath,amsthm,amssymb}
\usepackage{textcomp,booktabs}
\usepackage{footmisc}
\usepackage[normalem]{ulem}
\usepackage{multirow}
\usepackage{setspace}
\usepackage{bigstrut}
\usepackage{array}
\usepackage{amsfonts}
\usepackage{mathrsfs}
\usepackage{booktabs}
\usepackage{adjustbox}
\usepackage{caption}
\usepackage{cuted}
\usepackage{capt-of}
\usepackage{color, colortbl}
\usepackage[ruled,vlined]{algorithm2e}
\usepackage{amsmath}
\usepackage{amssymb}
\DeclareMathOperator*{\argmax}{argmax}
\usepackage{rotating}
\usepackage{commath}
\usepackage[ruled,vlined]{algorithm2e}
\usepackage{algorithmic}
\usepackage{dsfont}
\usepackage{adjustbox}
\usepackage{booktabs}

\ifCLASSOPTIONcompsoc
  \usepackage[nocompress]{cite}
\else
  \usepackage{cite}
\fi

\ifCLASSINFOpdf
\else
\fi

\newcommand{\x}{\boldsymbol{x}}

\newcommand{\Rb}{\mathbb{R}}

\newcommand{\Lm}{\mathcal{L}}
\newcommand{\etal}{\textit{et al.}}
\definecolor{Gray}{gray}{0.9}

\begin{document}

\title{Semantic Distribution-aware Contrastive Adaptation for Semantic Segmentation}
%
%
%
%

\author{Shuang Li, Binhui Xie, Bin Zang, Chi Harold Liu,~\IEEEmembership{Senior Member,~IEEE}, Xinjing Cheng, \\Ruigang Yang,~\IEEEmembership{Senior Member,~IEEE}, and Guoren Wang
\IEEEcompsocitemizethanks{ 
\IEEEcompsocthanksitem S. Li, B. Xie, B. Zang, C. H. Liu and G. Wang are with the School of Computer Science and Technology, Beijing Institute of Technology, Beijing, China. Corresponding author: C. H. Liu. Email: \{shuangli, binhuixie, binzang, chiliu, wanggrbit\}@bit.edu.cn\protect\\
\IEEEcompsocthanksitem X. Cheng and R. Yang are with Inceptio Technology, Shanghai, China. Email: cnorbot@gmail.com, ryang@cs.uky.edu \protect\\
}
\thanks{Manuscript received April 29, 2021.}
}

\markboth{SUBMITTED TO IEEE TRANSACTIONS ON PATTERN ANALYSIS AND MACHINE INTELLIGENCE}%
{Shell \MakeLowercase{\textit{et al.}}: Bare Advanced Demo of IEEEtran.cls for IEEE Computer Society Journals}
%


\IEEEtitleabstractindextext{%
\begin{abstract}
  Domain adaptive semantic segmentation refers to making accurate dense predictions on a certain target domain with only pixel-level annotations of a specific source domain. Current state-of-the-art works suggest that performing alignment from the category perspective can alleviate domain shift reasonably. However, they are mainly based on image-to-image adversarial training and little consideration is given to semantic variations of an object among different images. A possible consequence is that such alignment fails to capture a comprehensive picture of different categories, leading to unstable category alignment and limited generalization advances. This motivates us to explore a holistic representative, the semantic distribution from each category in the source domain, to mitigate the problem above. In this paper, we present a new \textit{semantic distribution-aware contrastive adaptation} algorithm, dubbed as SDCA, that enables pixel-wise representation alignment across domains under the guidance of the semantic distributions.
  To be precise, we first design a novel contrastive loss at pixel level by considering the correspondences between the semantic distributions and pixel-wise representations from both domains. Essentially, clusters of pixel representations from the same category are obliged to cluster together and those from different categories are obliged to spread out, boosting segmentation capability of the model. Next, an upper bound on this formulation is derived by implicitly involving the simultaneous learning of an infinite number of (dis)similar pixel pairs, making it highly efficient. Though simple, we empirically unveil the certain mechanisms that promote the potential of SDCA. Finally, we verify that SDCA can further improve the segmentation accuracy when integrated with the self-supervised learning method. We evaluate the proposed method on multiple benchmarks, 
  achieving considerable improvements over existing algorithms. 

\end{abstract}

\begin{IEEEkeywords}
Domain adaptation, semantic segmentation, semantic distribution, dense contrastive learning.
\end{IEEEkeywords}}

\maketitle

\IEEEdisplaynontitleabstractindextext

%
\IEEEpeerreviewmaketitle

\input{01-Introduction}
\input{02-Relatedwork}
\input{03-Method}

\input{04-Experiment}

\input{05-Analysis}

\input{06-Conclusion}

\ifCLASSOPTIONcaptionsoff
  \newpage
\fi

\bibliography{reference}
\bibliographystyle{IEEEtran}

\vspace{-10mm}
\begin{IEEEbiography}[{\includegraphics[width=1in,height=1.25in,clip,keepaspectratio]{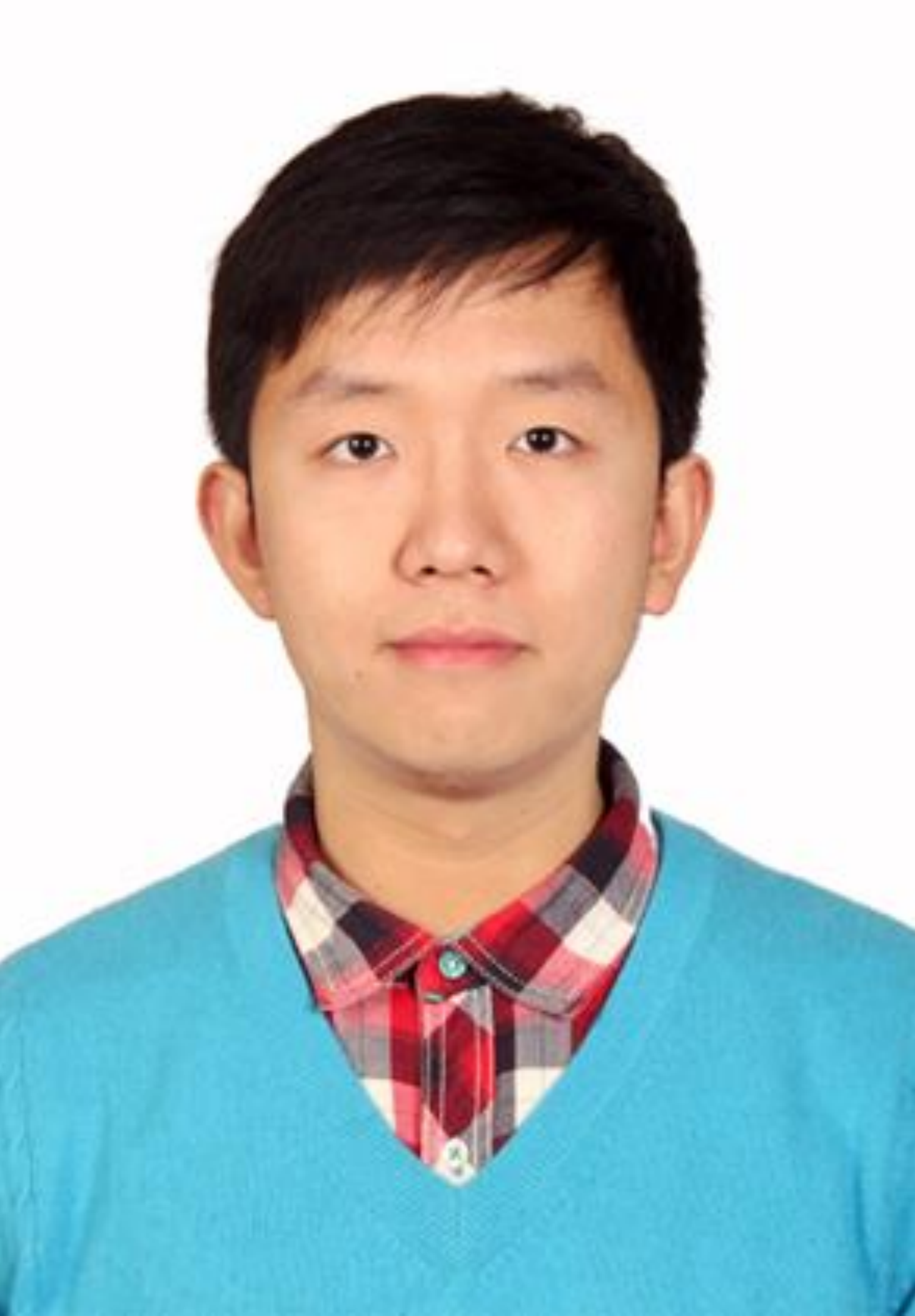}}]{Shuang Li} received the Ph.D. degree in control science and engineering from the Department of Automation, Tsinghua University, Beijing, China, in 2018.

He was a Visiting Research Scholar with the Department of Computer Science, Cornell University, Ithaca, NY, USA, from November 2015 to June 2016. He is currently an Assistant Professor with the school of Computer Science and Technology, Beijing Institute of Technology, Beijing. His main research interests include machine learning and deep learning, especially in transfer learning and domain adaptation.
\end{IEEEbiography}
\vspace{-10mm}

\begin{IEEEbiography}[{\includegraphics[width=1in,height=1.25in,clip,keepaspectratio]{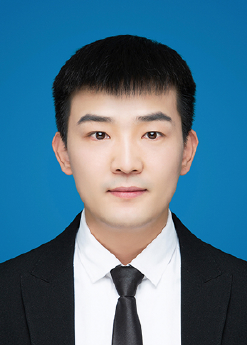}}]{Binhui Xie} is a graduate student at the School of Computer Science and Technology, Beijing Institution of Technology. His research interests focus on computer vision and transfer learning.
\end{IEEEbiography}
\vspace{-10mm}

\begin{IEEEbiography}[{\includegraphics[width=1in,height=1.25in,clip,keepaspectratio]{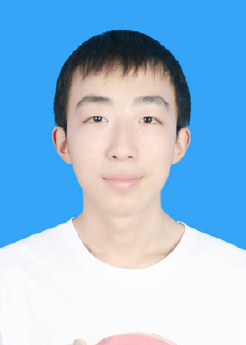}}]{Bin Zang} is a graduate student at the School of Computer Science and Technology, Beijing Institution of Technology. His research interests focus on computer vision and transfer learning.
\end{IEEEbiography}
\vspace{-10mm}

\begin{IEEEbiography}[{\includegraphics[width=1in,height=1.25in,clip,keepaspectratio]{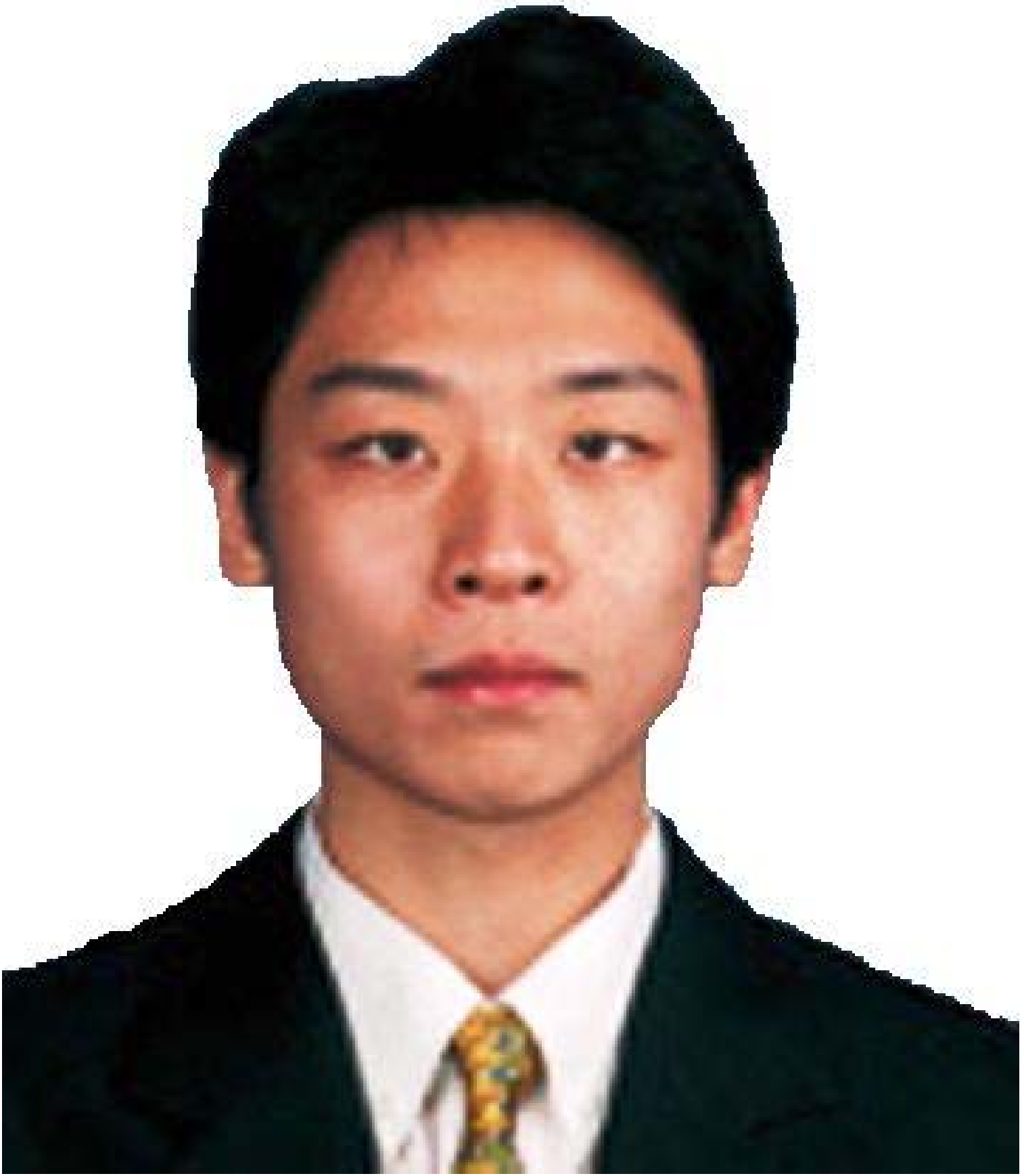}}]{Chi Harold Liu} receives the Ph.D. degree from Imperial College, UK in 2010, and the B.Eng. degree from Tsinghua University, China in 2006.

He is currently a Full Professor and Vice Dean at the School of Computer Science and Technology, Beijing Institute of Technology, China. Before moving to academia, he joined IBM Research - China as a staff researcher and project manager, after working as a postdoctoral researcher at Deutsche Telekom Laboratories, Germany, and a visiting scholar at IBM T. J. Watson Research Center, USA. His current research interests include the Big Data analytics, mobile computing, and deep learning. He has published more than 90 prestigious conference and journal papers and owned more than 14 EU/U.S./U.K./China patents. He is a Fellow of IET, and a Senior Member of IEEE.
\end{IEEEbiography}
\vspace{-10mm}

\begin{IEEEbiography}[{\includegraphics[width=1in,height=1.25in,clip,keepaspectratio]{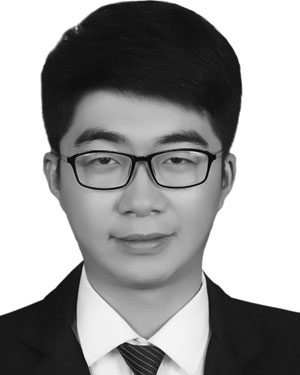}}]
{Xinjing Cheng}  is the head of perception team at Inceptio Tech., Shanghai, China. Before that, he was a research assistant with the Intelligent Bionic Center, Shenzhen Institutes of Advanced Technology (SIAT), Chinese Academy of Sciences(CAS), Shenzhen, China. His current research interests include computer vision, deep learning, robotics and autonomous driving.
\end{IEEEbiography}
\vspace{-10mm}

\begin{IEEEbiography}[{\includegraphics[width=1in,height=1.25in,clip,keepaspectratio]{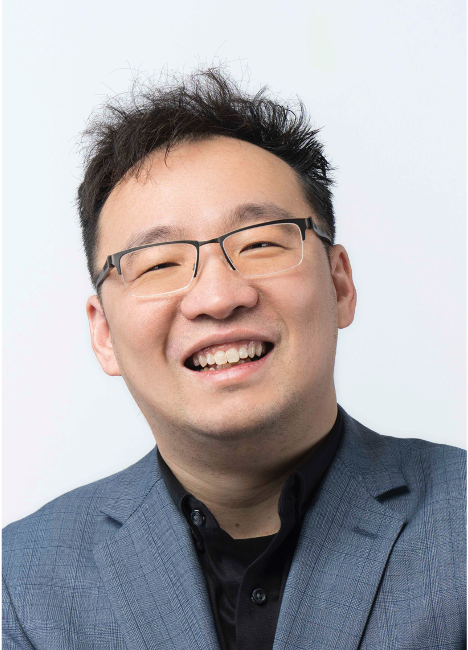}}]
  {Ruigang Yang} received his Ph.D. degree from University of North Carolina at Chapel Hill and M.S degree from Columbia University. He is the CTO of Inceptio and a full professor of computer science at the University of Kentucky. He was the director of Robotics and Autonomous Driving Lab at Baidu Research. He has published over 130 papers, which, according to Google Scholar, has receive over 14000 citations with an H-index of 61. He has received a number of awards, including US NSF Career award in 2004 and the Deans Research Award at the University of Kentucky in 2013.
\end{IEEEbiography}
\vspace{-10mm}

\begin{IEEEbiography}[{\includegraphics[width=1in,height=1.3in,clip,keepaspectratio]{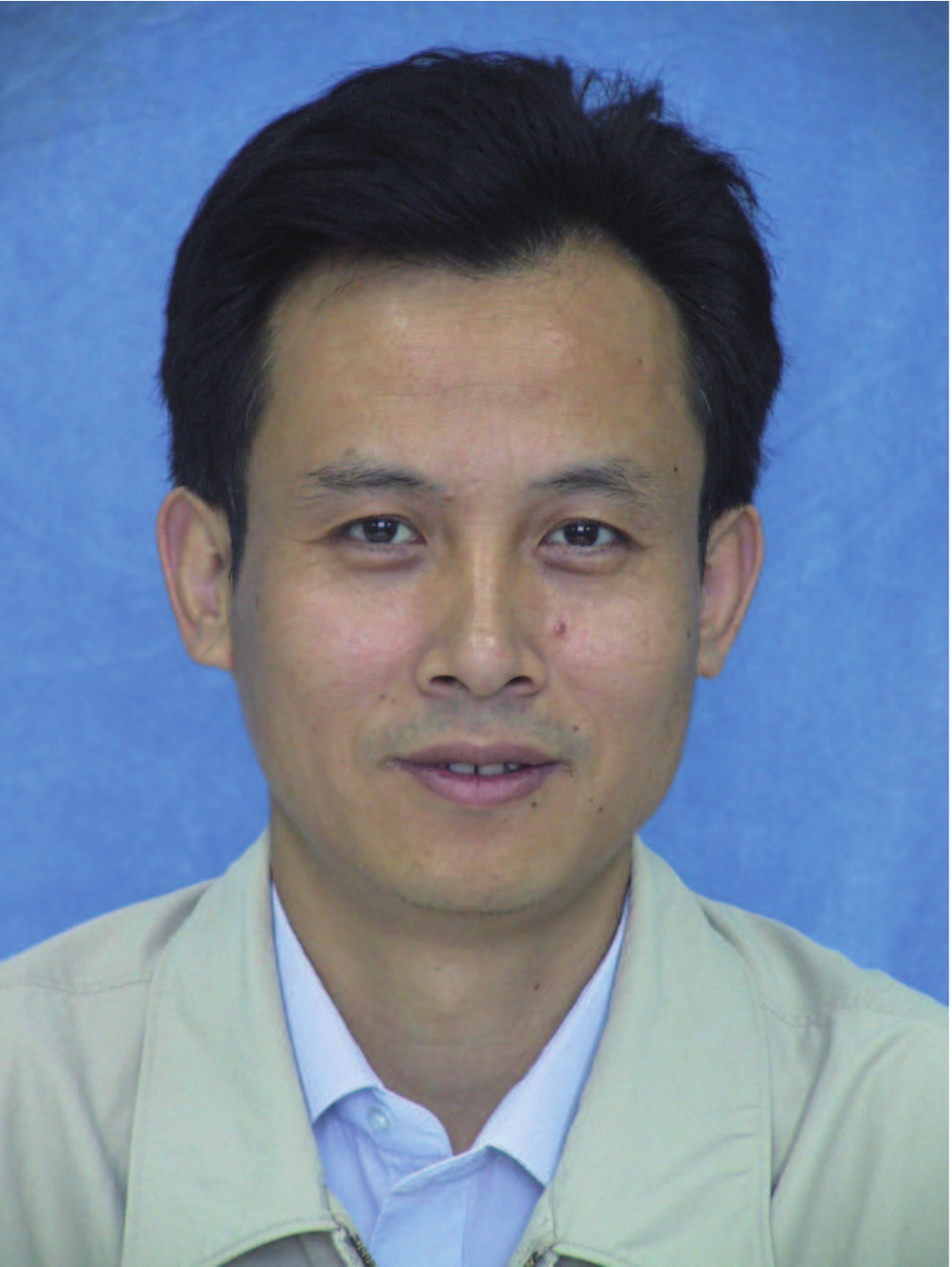}}]{Guoren Wang} received the BSc, MSc, and PhD degrees from the Department of Computer Science, Northeastern University, China, in 1988, 1991 and 1996, respectively. Currently, he is a Professor and the Dean with the School of Computer Science and Technology, Beijing Institute of Technology, Beijing, China. His research interests include XML data management, query processing and optimization, bioinformatics, high dimensional indexing, parallel database systems, and cloud data management. He has published more than 100 research papers.
\end{IEEEbiography}

\end{document}

%% file: 01-Introduction.tex
\ifCLASSOPTIONcompsoc
\IEEEraisesectionheading{\section{Introduction}}
\else
\section{Introduction}
\label{sec:introduction}
\fi

\IEEEPARstart{S}{emantic} segmentation is the task of assigning a semantic label to each pixel of a photograph. It is crucial to many downstream applications such as autonomous driving~\cite{geiger2012autonomous,zhou2020autonomous}, scene understanding~\cite{gupta2015scene,li2009scene}, and medical analysis~\cite{ronneberger2015UNet,mahapatra2013medical}. Albeit the rising of Convolution Neural Network (CNNs) has enabled us to make significant progress, deep CNNs are extremely data-hungry. As the labeling process for dense prediction tasks is expensive and labor intensive~\cite{Cordts2016Cityscapes}, recently, some works resort to synthetic images rendered from computer graphics~\cite{stephan2016gtav,ros2016synthia}. This is not the case, unfortunately, models trained on the simulated data often generalize poorly to realistic scenarios due to \textit{domain shift}~\cite{dataset_shift_in_ML09}. A semantic segmenter that can adapt itself to variations of conditions is always in demand. For instance, a segmenter used for autonomous driving is required to work well under diverse environments, i.e., a set of times, locations, and weather conditions. To tackle this challenge, domain adaptation (DA), which enables transferring a learning machine from a label-rich source domain to a label-scarce target domain, enjoys tremendous success in both vision~\cite{ouyang2021progressive,jiebo2018deep,luo2021category} and language~\cite{pan2010survey,liu2018corpus,liuli2019bow} communities. In this way, it could greatly relieve the need for laborious annotations and train a satisfactory segmenter. 

Researchers have developed adversarial training algorithms to minimize the domain discrepancies at the image level~\cite{dundar2020stylization,yang2020fda,li2019bidirectional}, feature level~\cite{hoffman2016fcns,luo2019significance,Hoffman_cycada2017}, or output level~\cite{tsai2018learning,vu2019advent,pan2020unsupervised}. To this end, the GAN-style architecture~\cite{goodfellow2014gan}, usually comprised of a generator and a discriminator, is widely used in this context. The generator extracts features from the raw images while the discriminator identifies the domain from which the features are generated. Therefore, the discriminator can guide the generator to extract target features with a distribution closer to the source feature distribution in an adversarial manner. Despite the above methods can draw the two domains closer through matching the marginal distributions, it does not guarantee that features from different categories in the target domain are well separated. This is because the feature distribution of each category should be described individually to take the semantic consistency into account. 

\begin{figure}[!htbp]
    \centering
    \includegraphics[width=0.48\textwidth]{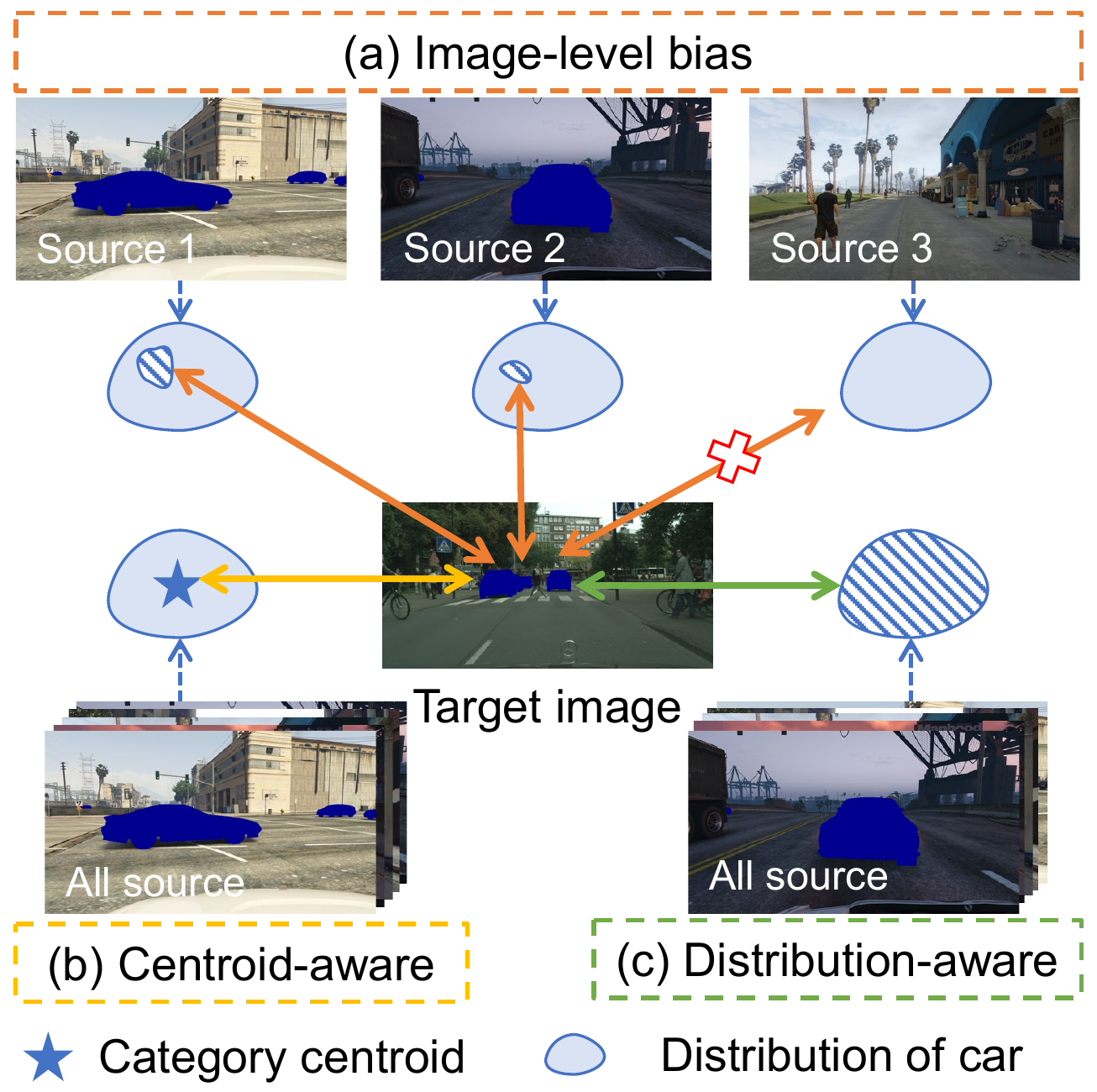}
    \caption{Illustration of the issues in previous category alignment approaches and our solution in the proposed SDCA. (a) {\bf Image-level bias}: during the process of alignment, the car instances vary across different source images due to imaging conditions, such as lighting and settings in the camera imaging pipeline. (b) {\bf Centroid-aware}: albeit category centroid can provide a holistic representative to guide the alignment, it might omit diversity, which impairs the discriminability of the learned features.  (c) {\bf Distribution-aware (Ours)}: towards pixel-wise discriminative learning, we seek semantic distribution in the same category as positive adaptation direction to pull them together and build some negative adaptation directions with the rest of semantic distributions to push them apart. This methodology is to learn pixel-wise representations that attract similar (positive) pixels and dispel different (negative) pixels, providing more clear decision boundary on the target domain.}
    \label{Fig_motivation}
    \vspace{-2mm}
\end{figure}

Recently, many approaches instead consider category alignment by incorporating semantic information with their features~\cite{chen2017nomorediscrimination,luo2021category,wang2020class,du2019ssf-dan,kang2020pixel}. These methods independently align semantic features across the source and target domains via category-level adversarial training. During adaptation, however, the mini-batch size used for segmentation is small (e.g., 1, as illustrated in Fig.~\ref{Fig_motivation} (a)) such that an object instance from source domain usually has large differences across images. In addition, it is common that some categories such as person and bike which appear frequently in the real-world images may disappear in a synthetic urban scene. Accordingly, these methods inevitably bring image-level bias, causing learned features prone to be unstably and wrongly aligned between two domains.

Although some existing works resort to category centroids computed on the source domain to lead the alignment~\cite{wang2020differential,zhang2019category}, yielding promising results, there still have some drawbacks. First, the centroid can represent the overall appearance of the category, but not the range of variations in certain attributes of the category, such as color, texture, and lighting. Hence, it might result in the degradation of the category diversity, which compromises the discriminative ability of the learned feature representations. Second, no attempt has been made in this regime to quantify the distance of different category features. It is likely hard to separate categories with similar distributions in the target domain especially when no supervision information is available, leading to results with large variance.

According to the analysis above, we approach domain adaptation in semantic segmentation from a new perspective and diminish domain shift via learning pixel-wise representations that attract similar pixels and dispel different pixels. First of all, we seek the distribution of each category in the source domain as the holistic representative to guide the directions of category alignment since the distribution can be properly estimated with sufficient supervision from source data. Our method is able to provide diverse generations from estimated distributions, thus covering the shortages of the category centroid based counterparts. 

Secondly, there is an insightful observation that increasing intra-category compactness and inter-category separability of pixel-wise representations can provide a better dense pixel classifier. Therefore, we separate pixel-wise representations in both source and target domains and implicitly define infinite number of positive pairs for each pixel by sampling from the estimated distribution in the same category. Meanwhile, infinite number of negative pixel pairs are drawn from the rest semantic distributions. Hereafter, a particular form of contrastive loss at the level of pixels is designed for contrastive adaptation. We further derive an upper bound on this formulation that makes it practically effective. 

Lastly, we propose to facilitate our framework with a self-supervised learning strategy, where we use segmentation predictions with high confidence to retrain the model and enhance the discriminability on the target domain. We conduct an analysis with Pixel-wise Discrimination Distance to certify the validity of our method regarding pixel-wise category alignment. Our experimental results have demonstrated that contrastively driving the source and target pixel-wise representations towards semantic distributions can effectively diminish the domain discrepancy and improve the generalization capability on the target domain. In a nutshell, our main contributions are summarized as follows.
\begin{itemize}
    \item We propose semantic distribution-aware contrastive adaptation (SDCA), a new learning algorithm for domain adaptive segmentation task. The essence is to explicitly encourage the connections between pixel-wise representations and semantic distribution of the same category as well as penalizing the connections between pixel-wise representations and semantic distributions of the different categories.
    
    \item An upper bound of the expected contrastive loss is derived with the statistics of the distribution in each category, which makes it simple yet effective to learn invariant and distinctive pixel-wise representations.
    \item Extensive empirical evaluations on several competitive benchmarks including SYNTHIA $\to$ Cityscapes, GTAV $\to$ Cityscapes, and Cityscapes $\to$ Cross-city demonstrate the SDCA improves the baseline model by significant margins. Analytical evidences are presented to validate its effectiveness.
\end{itemize}

%% file: 02-Relatedwork.tex
\section{Related Work}
\label{sec:relatedwork}
This section relates to a few research topics revolving around semantic segmentation, domain adaptation and contrastive learning. We clarify their similarities and differences compared to our method.

\subsection{Semantic Segmentation}
Semantic segmentation is a fundamental and challenging task in computer vision. The recent renaissance in this field began with fully convolutional networks~\cite{long2015fully}. Numerous methods have subsequently been studied to enlarge receptive fields and capture context information~\cite{yu2016dilated,zhao2017pspnet}. Among all these works, the family of Deeplab~\cite{chen2018deeplab,chen2018encoderdecoder,liu2019autodeeplab} enjoys remarkable effectiveness and simplicity that contribute to its magnetism and popularity. Unfortunately, these approaches require pixel-level annotations of large amounts of data, which stimulates interest in using synthetic data~\cite{stephan2016gtav,ros2016synthia}. In various practical applications, there is indeed a large gap between data distributions in train and test sets, which gives rise to severe performance loss at inference time. 
In this work, we deal with the semantic segmentation in the presence of domain shift that aims to learn only from source supervision a well performing model on target samples.

\subsection{Domain Adaptation}
Domain adaptation (DA) strives to alleviate domain shift and makes the knowledge learned from a similar but distinct source domain better transferred to a target domain of interest. Many pioneering works are proposed to tackle this problem in image classification tasks~\cite{tzeng2015simultaneous,DRCN,long2015dan,liuli2019guest,ouyang2018collaborative,long2016rtn,long2017jan}. Early domain adaptation works diminish the gap between domains by reweighting instances~\cite{huang2006correcting} or learning domain-invariant features~\cite{gong2012geodesic,DICD}. Subsequently, given the power of CNNs, various deep DA works have been discussed to boost transfer performance. A common strategy is to minimize the divergence between feature representations~\cite{long2015dan,long2017jan,tzeng2014deep}, e.g., maximum mean discrepancy~\cite{gretton2012mmd}. Drawn inspiration from generative adversarial network~\cite{goodfellow2014gan}, another popular line learns domain-invariant features via adversarial training~\cite{tzeng2017adversarial,ganin2015dann,long2018conditional,saito2018maximum}. Among this subset of methods, the most relevant works to ours are those learning semantic representations~\cite{kang2019contrastive,xie2018learning}. For instance, Kang~\etal~\cite{kang2019contrastive} explicitly model the intra-class and the inter-class discrepancies via optimizing a new metric named contrastive domain discrepancy.

\subsection{Domain Adaptation for Semantic Segmentation}
Unlike DA in image classification task, not until recently has the limited effort been made in semantic segmentation since its difficulty at dense structured predictions. Existing approaches can be categorized primarily into three groups: global alignment~\cite{tsai2018learning,dundar2020stylization,vu2019advent,yang2020fda,hoffman2016fcns}, category alignment~\cite{luo2021category,du2019ssf-dan,wang2020class,zhang2019category,wang2020differential,kang2020pixel}, and self-supervised learning~\cite{zou2018unsupervised,Zhang_2017_ICCV,zou2019confidence,zhang2020curriculum,li2019bidirectional}.

{\bf Global alignment.} Plenty of works fall into this group. Among them, adversarial training~\cite{goodfellow2014gan} is widely employed in two ways: (i) Mitigate the domain gaps from the image level via image style transfer across domains~\cite{Hoffman_cycada2017,dundar2020stylization}. Inspired by CycleGAN~\cite{CycleGAN2017}, Hoffman \etal~\cite{Hoffman_cycada2017} are among the first attempts to introduce DA to segmentation by building generative images for alignment. In~\cite{dundar2020stylization}, Dundar~\etal~propose to match the covariance of the universal feature embeddings across domains instead of carrying out expensive GAN training. These methods have proven that image-level adaptation remains important. (ii) Make the feature representations or the network predictions indistinguishable across domains~\cite{tsai2018learning,vu2019advent,pan2020unsupervised}. To name a few, Tsai \etal~\cite{tsai2018learning} suggest that performing alignment in the output space is more practical. Pan~\etal~\cite{pan2020unsupervised} propose to separate target domain into an easy and hard subdomain and minimize the intra-domain gap. Much works also align different properties between domains such as entropy~\cite{vu2019advent} and information~\cite{luo2021category}. Despite the fact that the marginal distribution discrepancy can be minimized by  global alignment, there is no guarantee that features from different categories in the target domain can be well separated.

{\bf Category alignment.} 
To enforce local semantic consistency on the target domain, Luo \etal~\cite{luo2019taking} propose to adaptively weight the adversarial loss while Du \etal~\cite{du2019ssf-dan} introduce multiple discriminators to separately adapt semantic features across domains. Similarly, Wang \etal~\cite{wang2020class} directly incorporate class information into the discriminator and encourage it to align features at a fine-grained level. 
However, due to the absence of holistic information about each category, the image-level bias is usually generated in the process of adversarial training. To address this issue, some methods instead adopt category centers computed on the source domain to guide the alignment between the two domains. For example, Zhang \etal~\cite{zhang2019category} propose a category-wise feature alignment guided by category anchors, which independently adapts semantic features across domains. Wang \etal~\cite{wang2020differential} treat the stuff regions and instances of things with different guidance during alignment. Further, Kang~\etal~\cite{kang2020pixel} put forward pixel associations to diminish the domain gap by contrastively strengthening their connections. 

In contrast, we endeavor to explore the category distribution as the comprehensive representative to enlarge the diversity of that category. On the other hand, we set forth a generic distribution-aware contrastive adaptation to emphasize pixel-wise discriminative learning, which allows us to minimize the intra-class discrepancy and maximize the inter-class margin of pixel features from both domains.

{\bf Self-supervised learning.} 
Another popular line, self-supervised learning, refers to giving pseudo labels to the unlabeled target samples, in which no human efforts are available~\cite{zou2018unsupervised,lian2019pycda,zou2019confidence,zhang2020curriculum}. In an example, Zou \etal~\cite{zou2018unsupervised} are among the first to propose an iterative learning strategy with class balance and spatial prior in the target domain. In~\cite{zhang2020curriculum}, Yang \etal~propose a curriculum-style learning approach to deal with easy tasks first and then infer necessary information about the target domain. Later on, Zou \etal~\cite{zou2019confidence} further propose confidence regularized self-training to prevent overconfident label belief. Most existing approaches established on pseudo labels which are noisy and heavily rely on a good initialization. Here, we demonstrate our pipeline is orthogonal to the self-supervised learning and performs favorably against previous approaches.

\subsection{Contrastive Learning}
To data, contrastive learning has been extensively investigated in unsupervised learning~\cite{hadsell2006dimensionality,oord2018infoNCE,he2020momentum,chen2020contrastive,cai2020jcl,chuang2020debiased}.  The central spirit of these methods aims to maximize the mutual information of latent representations among different augmented variations of an image. Very recently, some works~\cite{xie2020contrastive_dense,wang2020contrastive_dense} generalize contrastive learning to supervised dense prediction tasks. 
While the training recipes for standard contrastive learning have been highly mature and robust, the recipes for semantic segmentation are yet to be built, especially under the context of domain shift.

In this work, we focus on investigating several key factors for its success in domain adaptive semantic segmentation. 
Our method differs from them in the following ways. First, the optimization objectives are different. We aim to learn pixel-wise representations to distinguish different areas in an image for segmentation task instead of facilitating learning of meaningful image-wise representations for classification task. Second, our method considers the pixel-wise information for discriminative representation learning while previous methods target at instance separation. Finally, we novelly construct infinite similar/dissimilar pairs according to semantic distributions to diminish the domain discrepancy. This differs from conventional contrastive learning that defines finite image pairs via data augmentations.

%% file: 03-Method.tex
\section{Main Approach}
\label{sec:method}
In this section, we first introduce the problem definition and illustrate the overall idea in Section~\ref{sec:notation}. The details of our proposed approach are described in Section~\ref{sec:semantic_distribution_contrastive_adaptation}. Then we present sets of training objectives and the SDCA algorithm in Section~\ref{sec:objective}. Finally, in Section~\ref{sec:theory}, a profound theoretical insight of our method is provided.

\subsection{Problem Definition and Overall Idea}
\label{sec:notation}
\begin{figure*}[!htbp]
    \centering
    \includegraphics[width=0.96\textwidth]{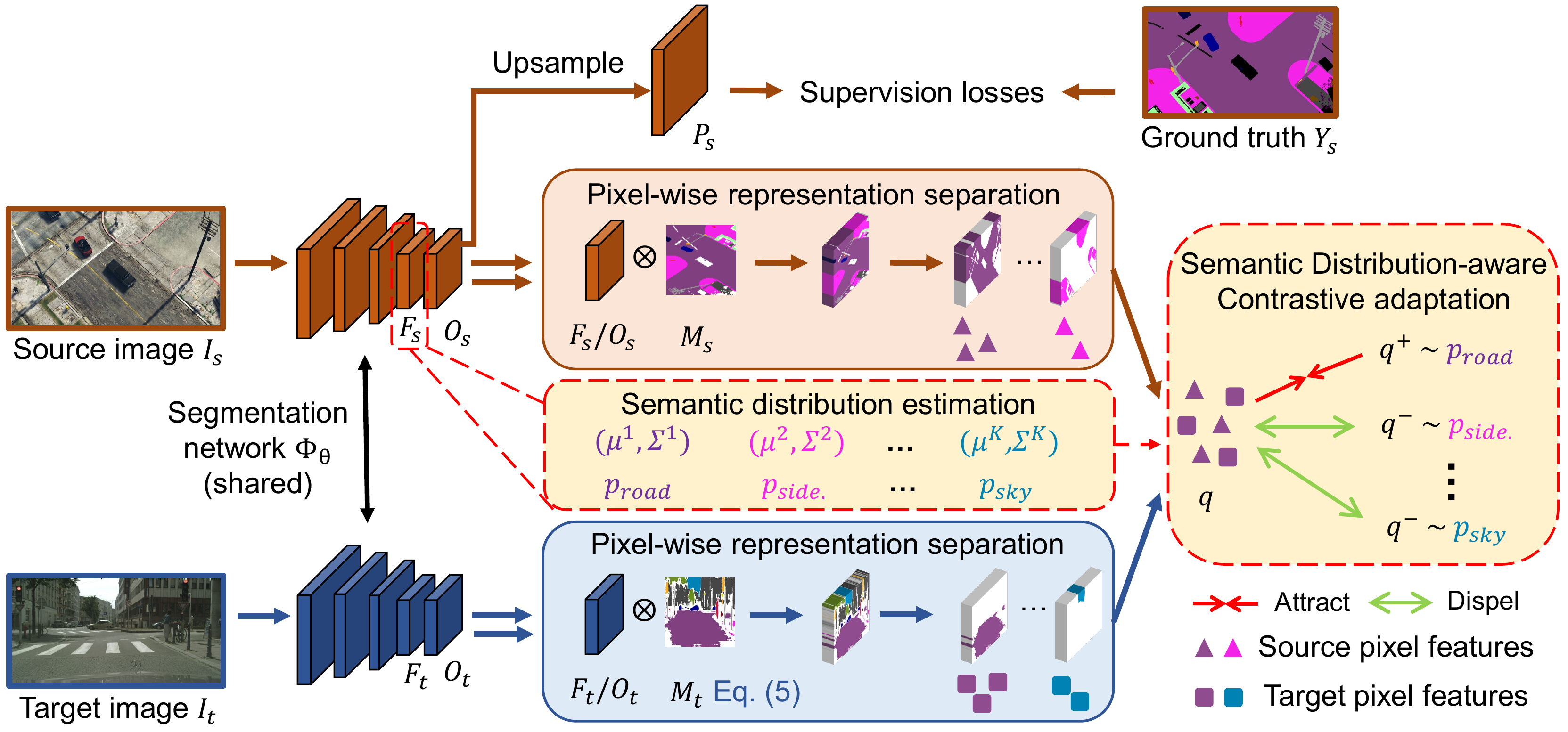}
    \caption{{\bf Framework of our SCDA.} It consists of three major components: (1) The \textit{segmentation network $\Phi_{\theta}$}, including a FCN-based feature generator and a multi-class classifier, maps the input image to high-level feature space and then transforms the feature space into the output space. (2) \textit{Semantic distribution estimation}, which guides the pixel-wise category alignment, dynamically estimates statistics of the distribution for each category on sufficient labeled source data. The mean of the features from a category represents the overall appearance and the covariance of each category captures intra-class variations. (3) \textit{Semantic distribution-aware contrastive adaptation} aims to learn pixel-wise discriminative representations. For each separated pixel representation from either source or target image, we have access to positive pixel pairs from the distribution that has the same semantic label and obtain negative pixel pairs from the rest distributions. With the designed contrastive loss, we can encourage similar pixel representations to be close, and those of dissimilar to be
    more orthogonal.}
    \label{Fig_framework}
    \vspace{-3mm}
\end{figure*}

In domain adaptive semantic segmentation, we have a collection of labeled source data $\mathcal{S}$ as well as unlabeled target data $\mathcal{T}$. The goal is to train a segmentation network $\Phi_{\theta}$ that categorizes each pixel of a target image into one of the predefined $K$ classes. During training, the network $\Phi_{\theta}$ is trained with labeled source data and unlabeled target data.
Images from the source and target domains $I_s\,, I_t\in \Rb^{H\times W\times 3}$ are randomly sampled and passed into the network. Then, we can acquire their features $F_s\,, F_t\in \Rb^{H'\times W'\times A}$, segmentation outputs $O_s\,, O_t\in \Rb^{H'\times W'\times A}$, and upsampled pixel-level predictions $P_s\,, P_t\in \Rb^{H\times W\times K}$. $A$ is the channel dimension of intermediate features and $H' (\ll H)\,, W' (\ll W)$ are downsampled spatial dimensions of an image after passing through the network as illustrated in Fig.~\ref{Fig_framework}. 

The key idea underlying our approach is to diminish the domain gap via learning pixel-wise discriminative representations under the guidance of semantic distributions. To implement this idea, a straightforward way is to directly adopt category anchors computed on the source domain to guide the alignment between both domains. The problem of this design is that it does not fully unleash the potential strength of category information, and anchors only reflect the common characteristic of each category, which results in erroneous representation learning. Our experimental results have demonstrated that semantic distributions can be represented as more informative guidance, increasing the robustness.
On the other hand, we contrastively strengthen the connections between each pixel representation and the estimated semantic distributions, which adjusts the relative magnitude of inter-class and intra-class feature distances. 

\subsection{Semantic Distribution-aware Contrastive Adaptation}
\label{sec:semantic_distribution_contrastive_adaptation}
The pipeline of our SDCA is intuitive: first we exploit the statistics of the distribution for each category to mine comprehensive semantic information (Section~\ref{sec:sematin_distribution}); then, we design a novel contrastive loss that involves the simultaneous learning of an infinite number of similar/dissimilar pairs at pixel level to mitigate the domain gap (Section~\ref{sec:contrative_adaptation}); besides applying contrastive adaptation to the penultimate feature maps, i.e., $F_s\,, F_t$, we also apply adaptation on the final network outputs, i.e., $O_s\,, O_t$ (Section~\ref{sec:multi-level}).

\subsubsection{Sketch of Contrastive Learning}
Contrastive learning~\cite{hadsell2006dimensionality,he2020momentum,chen2020contrastive,cai2020jcl} has recently shown to be a highly effective way to learn meaningful representations from unlabeled data. Let $f$ be an embedding function (realized via a CNN) that transforms an sample $x$ to an embedding vector $z=f(x)\,, z \in \mathbb{R}^d$. Then, we normalize $z$ onto a unit sphere. Let $(x\,,x^+)$ be similar pairs and $(x\,,x^-)$ be dissimilar pairs. A popular contrastive loss such as InfoNCE~\cite{oord2018infoNCE} is formulated as:
\begin{small}
    \begin{align}
        \mathbb{E}_{x\,,x^+\,,\{x^{n-}\}_{n=1}^N} \left[- \log \frac{e^{f(x)^{\top}f(x^+)/\tau}}{e^{f(x)^{\top}f(x^+)/\tau} + \sum_{n=1}^N e^{f(x)^{\top}f(x^{n-})/\tau}}\right].
        \label{eq:infonce}
    \end{align}
\end{small}
In practice, the expectation is replaced by the empirical estimate. As shown above, the contrastive loss is essentially based on the softmax formulation with a temperature $\tau$~\cite{oord2018infoNCE}. 

{\bf Discussion.} Intuitively, the above contrastive loss encourages the instance discrimination. On the contrary, our work explores dense pixel predictions for DA in semantic segmentation, which have received limited consideration in prior researches. We show that pixel-wise representation alignment surpasses the existing algorithms by a significant margin, demonstrating the potential of this direction.

\subsubsection{Semantic Distribution Estimation}
\label{sec:sematin_distribution}
Given the source feature map $F_s \in \Rb^{H'\times W'\times A}$, for an arbitrary source pixel $i\in \{1\,,2\,,\cdots\,, H'\times W'\}$ in $F_s$, we first separate the pixel-wise representations via the downsampled ground truth label, i.e., source mask $M_s \in \Rb^{H'\times W'}$. Then, the mean of features from the $k^{th}$ category is calculated as the average values of every single dimension in the feature vector,
\begin{small}
    \begin{align}
        \mu'^k = \frac{1}{|\Lambda^k|} \sum_{i\in \{1\,,2\,,\cdots\,, H'\times W'\}} \mathds{1}_{[M_{s, i}=k]} F_{s, i}\,,
        \label{eq:semantic_prototype_initialize}
    \end{align}
\end{small}%
where $F_{s,i}\in \Rb^{A}$ is feature representation of source pixel $i$ and $\mathds{1}$ is an indicator function which returns 1 if the condition holds or 0 otherwise. $\Lambda^k$ denotes the pixel set that contains all the pixel features belonging to the $k^{th}$ semantic class within $F_{s}$ and $|\cdot|$ is the number of pixels in the set. 

In implementation, such calculation is computational-intensive on the whole source domain. To address this, the mean is computed in an online fashion by aggregating statistics one by one. Mathematically, the online estimation algorithm for mean is given by:
\begin{small}
    \begin{align}
        \mu^k_{(t)} = \frac{n^k_{(t-1)}\mu^k_{(t-1)} + m^k_{(t)}{\mu'}^k_{(t)}}{n^k_{(t-1)}+m^k_{(t)}}\,,
        \label{eq:mu}
    \end{align}
\end{small}%
where $n^k_{(t-1)}$ represents the total number of pixels belonging to $k^{th}$ category in previous $t-1$ images, and $m^k_{(t)}$ represents the number of pixels belonging to $k^{th}$ category only in $t^{th}$ image.
As the feature vector $F_{s,i}$ is multi-dimensional, we use covariance for a better representation of the variance between any pair of elements in the feature vector. The covariance matrix $\Sigma^k$ for class $k$ is calculated as:
\begin{small}
    \begin{align}
        \Sigma^k_{(t)} &= \frac{n^k_{(t-1)}\Sigma^k_{(t-1)}+m^k_{(t)}{\Sigma'}^k_{(t)}}{n^k_{(t-1)}+m^k_{(t)}} \nonumber\\
        &+\frac{n^k_{(t-1)}m^k_{(t)}\left(\mu^k_{(t-1)}-{\mu'}^k_{(t)}\right)\left(\mu^k_{(t-1)}-{\mu'}^k_{(t)}\right)^{\top}}{\left(n^k_{(t-1)} + m^k_{(t)}\right)^2} \,,
        \label{eq:Sigma}
    \end{align}%
\end{small}%
where ${\Sigma'}^k_{(t)}$ denotes the covariance matrix of the features of the $k^{th}$ category in $t^{th}$ image.
It is noteworthy that before training, $K$ mean values and $K$ covariance matrices are computed on the whole source domain as the initialization, one for each category. During adaptation, we dynamically update these semantic distributions for each given source training image. The estimated semantic distributions are more informative for guiding the category alignment. 

\subsubsection{Contrastive Adaptation}
\label{sec:contrative_adaptation}
Recently, several prior methods have leveraged category feature centroids~\cite{zhang2019category} or instance and stuff features~\cite{wang2020differential} in the source domain serve as anchors to remedy the domain shift problem. However, in their works, these anchors merely preserve the basic characteristic of each category, but at the expense of the diversity within the category. Additionally, the margin between the categories are not explicitly enlarged, which severely limits their potential capability in dense prediction tasks. 

On the contrary, our method is different from previous methods in that it fully exploits the statistics of the distribution to guide the category alignment at the level of pixels. In our framework, we derive a particular form of the contrastive loss where multiple positive/negative pixel pairs are simultaneously involved with regard to each pixel representation in the source and target domain. The central spirit of this modification is to force various similar/dissimilar pairs to build up stronger intra-/inter-category connections for domain adaptive semantic segmentation tasks. 

As described above, the pixel-wise representation separation is naturally guaranteed with source mask $M_s$. However, for target domain data, the most important issue is how to obtain satisfied target mask $M_t$ for each pixel, since the training error could be amplified due to the noisy predictions when generating the target mask. 
To remedy this, we employ a confident strategy with a confidence threshold $\delta$. In detail, a confidence map is generated according to the segmentation prediction map $O_t$, where the confidence value is the maximum item of the softmax output in each channel. This enables the mask at each pixel to be associated with a confidence value, i.e., maximum prediction probability. 
With the $\delta$ being set, we can define target mask as follows:
\begin{small}
    \begin{align}
        M_{t,j} = \mathop{\arg\max}\limits_{k} O_{t,j}^k\,,    \text{subject to}~O_{t,j}^k\ge\delta\,.
        \label{eq:target_mask}
    \end{align}   
\end{small}%
To this end, every pixel representation in the source and target features (for simplicity we define it as query $q_i\in \Rb^{A}$) now needs to return a low loss value when simultaneously paired with multiple positive pixel pairs $q^{m+}$ and negative pixel pairs $q^{n-}_j$, where $q^{m+}$ indicates the $m^{th}$ positive example from the same category w.r.t. $q_i$ and $q^{n-}_j$ represents $n^{th}$ negative example from the $j^{th}$ different category. Formally, we define a novel pixel-wise contrastive loss as: 
\begin{small}
    \begin{align}
        \mathcal{L}^{M,N}_i = -\frac{1}{M}\sum_{m=1}^M\log\frac{e^{q_i^{\top} q^{m+}/\tau}}{e^{q_i^\top q^{m+}/\tau} + \sum_{j=1}^{K-1}\frac{1}{N}\sum_{n=1}^N e^{q_i^\top q^{n-}_{j}/\tau}}\,,
    \end{align}
\end{small}
where $M$ and $N$ are the numbers of positive and negative examples respectively.
A naive implementation of $\mathcal{L}^{M,N}$ is to explicitly sample $M$ examples from semantic distribution that has the same latent class and $N$ examples from each of the other distributions that has different semantic label. Unfortunately, this is not computational applicable when $M$ and $N$ is large, as carrying all positive/negative pixel pairs in an iteration would quickly drain the GPU memory. 

To get around this problem, we take an infinity limit on the number of $M$ and $N$, where the effect of $M$ and $N$ is hopefully absorbed in a probabilistic way. With this application of infinity limit, the statistics of the data are sufficient to achieve the same goal of multiple pairing. Mathematically, as $M$ and $N$ goes to infinity, $\mathcal{L}^{M,N}$ becomes the estimation of:
\begin{small}
    \begin{align}
        \mathcal{L}^{\infty}_i &= \lim_{\substack{M\rightarrow \infty\\N\rightarrow \infty}} \mathcal{L}^{M,N}_i \nonumber\\
        & = -\mathbb{E}_{\substack{q^{+} \sim p(q^+) \\ q^{-}_j\sim p(q^{-}_j)}} \log\frac{e^{q_i^\top q^+/\tau}}{e^{q_i^\top q^+/\tau} + \sum_{j=1}^{K-1} e^{q_i^\top q^{^-}_j/\tau}} \,,
        \label{eq:infinite_loss}
        \end{align}
\end{small}
where $p(q^+)$ is the positive semantic distribution that has the same semantic label and $p(q^{-}_j)$ is the $j^{th}$ negative semantic distribution that has different semantic label with respect to $q_i$. The analytic form of Eq.~\eqref{eq:infinite_loss} itself is intractable, but Eq.~\eqref{eq:infinite_loss} has a rigorous closed form of upper bound, which can be derived as:
\begin{small}
    \begin{align}
    &-\mathbb{E}_{\substack{q^{+}, q^{-}_j}} \log\frac{e^{q_i^\top q^+/\tau}}{e^{q_i^\top q^+/\tau} + \sum_{j=1}^{K-1} e^{q_i^\top q^{-}_j/\tau}} \nonumber\\
    &= \mathbb{E}_{q^+}\left[ \log \left[ e^{\frac{q_i^\top q^+}{\tau}} + \sum_{j=1}^{K-1} \mathbb{E}_{q^{-}_j}e^{\frac{q_i^\top q^{-}_j}{\tau}}\right]\right] - \mathbb{E}_{q^+}\left[ \frac{q_i^\top q^+}{\tau}\right] \\
    &\leq \log\left[ \mathbb{E}_{q^+}\left[e^{\frac{q_i^\top q^+}{\tau}}+\sum_{j=1}^{K-1} \mathbb{E}_{q^{-}_j}e^{\frac{q_i^\top q^{-}_j}{\tau}}\right]\right] - q_i^\top\mathbb{E}_{q^+}\left[ \frac{q^+}{\tau}\right] \label{eq:jensen} \\
    &= \log\left[\mathbb{E}_{q^+}e^{\frac{q_i^\top q^+}{\tau}} + \sum_{j=1}^{K-1} \mathbb{E}_{q^{-}_j}e^{\frac{q_i^\top q^{-}_j}{\tau}}\right] - q_i^\top\mathbb{E}_{q^+}\left[\frac{q^+}{\tau}\right] \label{eq:upper_bound} \\
    &= \Bar{\mathcal{L}_i} 
    \end{align}
\end{small}
where the inequality Eq.~\eqref{eq:jensen} follows form the Jensen's inequality on concave functions, i.e., $\mathbb{E} \log(X) \leq \log \mathbb{E}\left[ X \right]$. 
To facilitate our formulation, we need some further assumptions on the feature distribution. Specifically, we assume that $q^{+} \sim \mathcal{N}(\mu^{+}, \Sigma^{+})$ and $q^{-}_j\sim \mathcal{N}(\mu^{-}_j, \Sigma^{-}_j)$, where $\mu^{+}$ and $\Sigma^{+}$ are respectively the statistics i.e., mean and covariance matrix, of the positive semantic distribution for $q$, $\mu^{-}_j$ and $\Sigma^{-}_j$ are respectively the statistics of the $j^{th}$ negative distribution.

For any random variable $x$ that follows Gaussian distribution $x\sim \mathcal{N}(\mu, \Sigma)$, where $\mu$ is the expectation of $x$, $\Sigma$ is the covariance matrices of $x$, we have the moment generation function~\cite{wang2021isda} that satisfies:
\begin{small}
    \begin{align}
        \mathbb{E} \left[e^{a^\top x}\right] = e^{a^\top \mu + \frac{1}{2}a^\top \Sigma a}\,.
        \label{eq:moment_generation_function}
    \end{align}%
\end{small}%

Under the Gaussian assumption $q^{+} \sim \mathcal{N}(\mu^{+}, \Sigma^{+})\,, q^{-}_j\sim \mathcal{N}(\mu^{-}_j, \Sigma^{-}_j)$, along with Eq.~\eqref{eq:moment_generation_function}, we find that Eq.~\eqref{eq:upper_bound} for a certain pixel representation $q_i$ immediately reduces to:
\begin{small}
    \begin{align}
    &\Bar{\mathcal{L}_i} \nonumber\\
    &=\log \left[e^{\frac{q_i^\top \mu^{+}}{\tau} + \frac{q_i^{\top}\Sigma^{+} q_i}{2\tau^2}} + \sum_{j=1}^{K-1} e^{\frac{q_i^\top \mu_{j}^{-}}{\tau} + \frac{q_i^\top \Sigma_{j}^{-} q_i}{2\tau^2}} \right] - \frac{q_i^\top\mu^{+}}{\tau} \\
    &=-\log\frac{e^{\frac{q_i^\top \mu^{+}}{\tau}+\frac{q_i^\top \Sigma^{+} q_i}{2\tau^2}}}{e^{\frac{q_i^\top\mu^{+}}{\tau}+\frac{q_i^\top\Sigma^{+} q_i}{2\tau^2}}+\sum_{j=1}^{K-1} e^{\frac{q_i^\top\mu_{j}^{-}}{\tau}+\frac{q_i^\top\Sigma_{j}^{-} q_i}{2\tau^2}}} + \frac{q_i^\top\Sigma^{+}q_i}{2 \tau^2} \,.
    \end{align}
\end{small}
The overall loss function with regard to each feature map in the source and target domain thereby boils down to the closed form whose gradients can be analytically solved for:
\begin{small}
    \begin{align}
        \mathcal{L}_{feat} = \frac{1}{|F_s|} \sum_{i\in F_s} \Bar{\mathcal{L}_i}  + \frac{1}{|F_t|} \sum_{j\in F_t} \Bar{\mathcal{L}_j} \,,
        \label{eq:feat_contrastive}
    \end{align}%
\end{small}%
where $|F_s|$ and $|F_t|$ are respectively numbers of pixels in $F_s$ and $F_t$.
Based on this, an effective semantic distribution-aware contrastive loss is yielded to mitigate domain discrepancy via learning discriminative pixel representations. 

Note that the proposed contrastive loss is employed in both domains simultaneously. For one thing, when the loss is applied in the source domain, the network $\Phi_{\theta}$ is able to learn more discriminative feature representations for pixel-level predictions, which could increase the robustness of the segmentation model. Another effect brought by Eq.~\eqref{eq:feat_contrastive} is that the target pixel-wise representations are contrastively adapted, which benefits minimizing the intra-category discrepancy and maximizing the inter-category margin on the target domain and facilitates transferring knowledge from source to target explicitly.

\subsubsection{Multi-Level Contrastive Adaptation}
\label{sec:multi-level}
Aside from applying contrastive adaptation to the penultimate feature maps of the network, we also employ adaptation on the segmentation outputs. Such multi-level adaptation has been proved to be able to further eliminate the domain shift~\cite{kang2020pixel,tsai2018learning,luo2019taking}. 

We follow the same rule as discussed in Section~\ref{sec:contrative_adaptation} to perform the adaptation on the output space, i.e.,
$q_i\in O_{s}\cup O_{t}$. Similarly, the loss function with regard to each segmentation output in the two domains are given by:
\begin{small}
    \begin{align}
        \mathcal{L}_{out} = \frac{1}{|O_s|} \sum_{i\in O_s} \Bar{\mathcal{L}_i}  + \frac{1}{|O_t|} \sum_{j\in O_t} \Bar{\mathcal{L}_j} \,.
        \label{eq:out_contrastive}
    \end{align}%
\end{small}%
Applying multi-level contrastive adaptation not only reduces the effect of overfitting to the source domain, but also mines and exploits multi-granular pixel-wise relationships with estimated semantic distributions. The multi-level adaptations are complementary to each other and both of them contribute to alleviating the domain shift.

\subsection{Training Objective}
\label{sec:objective}
Regarding the training objectives, we train the network $\Phi_{\theta}$ with standard segmentation supervision loss, i.e., pixel-wise cross-entropy loss, for the source domain image:  
\begin{small}
    \begin{align}
        \Lm_{ce} = -\frac{1}{|I_s|}\sum_{i\in I_s} \sum_{k} \mathds{1}_{[Y_{s,i} = k]} \log P_{s,i}^{k}\,,
        \label{eq:ce_loss}
    \end{align}%
\end{small}
where $I_s$ denotes the source image with its ground truth label $Y_s$.
Moreover, to improve segmentation quality, in particular on the tail classes (ref to Fig.~\ref{Fig_tail_class}),  the Lov{\'{a}}sz-Softmax~\cite{berman2018lovasz} loss $\Lm_{lov}$ is imposed on the source data. Since $\Lm_{lov}$ is a convex surrogate to the Jaccard loss, it directly optimizes the intersection-over-union of each instance. Therefore we can slightly alleviate the negative effect of category imbalance.

Meanwhile, we employ the multi-level contrastive adaptation to mitigate the domain shift in conjunction with the above loss functions. The overall training objective is formulated as follows:
\begin{small}
    \begin{align}
        \mathop{\min}\limits_{\Phi_{\theta}} \Lm_{ce} + \lambda_{lov} \Lm_{lov} + \lambda_{feat} \Lm_{feat} + \lambda_{out} \Lm_{out}\,,
        \label{eq:overall_loss}
    \end{align}
\end{small}
where $\lambda_{lov}\,,\lambda_{feat}\,,\lambda_{out}$ are constants controlling the strength of corresponding loss. Initial tests suggest that using equal weights to combine the the $\Lm_{feat}$ with $\Lm_{out}$ yields better results. For simplicity, $\lambda_{feat}$ and $\lambda_{out}$ are set to 1.0 without any tuning. 
By optimizing Eq.~\eqref{eq:overall_loss}, clusters of pixels belonging to the same category are pulled together in the feature space while synchronously pushing apart from other categories. In this way, our method can simultaneously minimize the domain gap across domains as well as enhancing the intra-class compactness and inter-class separability in a unified framework. Here, we summarize our training process in Algorithm~\ref{alg:contrastive}.

As we all know, self-supervised learning has been used in previous methods~\cite{zhang2019category,wang2020differential,wang2020class,pan2020unsupervised,li2019bidirectional}, which is usually achieved by iteratively generating a set of pseudo labels based on the most confident predictions on the target data. Unfortunately, it primarily depends on a good initialization model and is hard to tune. Our contrastive adaptation strategy is over the feature representations and is orthogonal to the self-supervised learning. Therefore, we propose a self-supervised learning framework combined with our contrastive adaptation to further promote the generalization ability of the model. 

Specifically, once the alignment is finished, we can generate the reliable pseudo labels for target data. Different from assigning target masks based on the constant threshold $\delta$, we pre-define the median threshold $\delta^k$ according to the output predictions over category $k$, meaning 50\% pixels in the category $k$ have confidence above $\delta^k$. The target pseudo labels in the output space are obtained: 
\begin{small}
    \begin{align}
        \hat{Y}_{t,j} = \mathop{\argmax}\limits_{k} P_{t,j}^k \,, \text{subject to}~P_{t,j}^k \ge\delta^k\,.
        \label{eq:target_pseudo_label}
    \end{align}
\end{small}

Then, we fine-tune the model via optimizing a self-supervised loss $\Lm_{ssl}$ in Eq.~\eqref{eq:self_training} on the entire target training data to make the model more adaptive to the target domain.
\begin{small}
    \begin{align}
        \mathop{\min}\limits_{\Phi_{\theta}} -\frac{1}{|I_t|}\sum_{j\in I_t} \sum_{k} \mathds{1}_{[\hat{Y}_{t,j} = k]} \log P_{t,j}^{k} \,.
        \label{eq:self_training}
    \end{align}
\end{small}%

\begin{algorithm}[htbp]
        \SetAlgoLined
        \KwIn{\\
        1) The ImageNet pre-trained DeepLab-v2 netwrok $\Phi_{\theta}$.\\
        2) The source domain $\mathcal{S}$ and target domain $\mathcal{T}$. \\
        3) The confident threshold $\delta$, maximum iterations $L$, and hyper-parameters $\lambda_{lov}\,, \lambda_{feat}\,, \lambda_{out}$.
        
        \nl Warm-up model $\Phi_{\theta}$ according to Eq.~\eqref{eq:ce_loss}.
        }

        \nl Initialize statistics $\{\mu^{k}\}_{k=1}^{K}$ and $\{\Sigma^{k}\}_{k=1}^{K}$ using $\mathcal{S}$.

        \nl \For{l $\leftarrow$ 0 to $L$}{
            \nl Randomly sample a source image $I_{s}$ with $Y_{s}$ from $\mathcal{S}$ and a target image $I_{t}$ from $\mathcal{T}$.

            \nl Compute the feature maps $F_{s}\,, F_{t}$, segmentation outputs $O_{s}\,, O_{t}$ and pixel-level prediction $P_{s}$.

            \nl Estimate current mean values $\{\mu^{k}_{(t)}\}_{k=1}^{K}$ via Eq.~\eqref{eq:mu} and covariance matrices $\{\Sigma^{k}_{(t)}\}_{k=1}^{K}$ via Eq.~\eqref{eq:Sigma}.

            \nl Separate pixel-wise representations of both domains in the feature space and output space according to their masks $M_{s}$ and $M_{t}$.

            \nl Train $\Phi_{\theta}$ using losses $\Lm_{ce}\,,\Lm_{lov}\,, \Lm_{feat}$ and $\Lm_{out}$.
        }
        \Return{$\Phi_{\theta}$}
        \caption{{\bf SDCA algorithm.} \label{alg:contrastive}}
\end{algorithm}

\subsection{Theoretical Insight}
\label{sec:theory}
Motivated by the theory of domain adaptation in~\cite{david2010theory}, we provide a theoretical insight to investigate why our framework can generalize well on the target domain. 
Let $h \in \mathcal{H}$ be a hypothesis and denote by $\epsilon_s(h)$ and $\epsilon_t(h)$ the expected risks of source and target domain respectively, then
\begin{small}
    \begin{align}\label{eq-target-error-bound}
        \begin{split}
        \epsilon_t(h)&\leq \epsilon_s(h)+\frac{1}{2}d_{\mathcal{H}\Delta\mathcal{H}}(\mathcal{S},\mathcal{T})+\varsigma  \,,\\
        \end{split}
    \end{align}%
\end{small}%
where
\begin{small}
    \begin{align}\label{eq-Dis-HdeltaH}
        \notag
        &d_{\mathcal{H}\Delta\mathcal{H}}(\mathcal{S},\mathcal{T})\\
        \notag
        &\triangleq 2\mathop{\sup}\limits_{h,h'\in \mathcal{H}}\left | \mathop{\mathbf{Pr}}\limits_{\x\thicksim \mathcal{S}}\left[h(\x)\neq h'(\x)\right]-\mathop{\mathbf{Pr}}\limits_{\x\thicksim \mathcal{T}}\left[h(\x)\neq h'(\x)\right] \right | \,,\\
        &\varsigma  \triangleq \mathop{min}\limits_{h \in \mathcal{H}} \left\{\epsilon_s(h, h_{\mathcal{S}}) + \epsilon_t(h, h_{\mathcal{T}})\right\} \,,
    \end{align}%
\end{small}%
where $h_{\mathcal{S}}$ and $h_{\mathcal{T}}$ are labeling functions for the source and target domain respectively. In practice, $e_s(h)$ is the expected error on the source domain which can be minimized easily with the guidance of source ground truth annotations. 

Regarding to $d_{\mathcal{H}\Delta\mathcal{H}}(\mathcal{S},\mathcal{T})$, it actually represents the divergence between the distributions $\mathcal{S}$ and $\mathcal{T}$. The proposed contrastive adaptation aims at optimizing $d_{\mathcal{H}\Delta\mathcal{H}}(\mathcal{S},\mathcal{T})$ from local joint distribution perspective. Recall that in Eq.~\eqref{eq:infinite_loss}, we hope to leverage a distribution-aware contrastive loss to minimize the intra-category discrepancy, i.e., the domain shift within the same category, and maximize the inter-category margin, i.e., the domain shift among different categories, boosting category alignment at pixel representation level. We empirically verify that this term is decreased in Section~\ref{sec:analysis}. 

As for the third term $\varsigma$, recall the triangle inequality which suggests that for any labeling function $h_1\,, h_2\,, h_3 \in \mathcal{H}$, we have $\epsilon(h_1\,, h_2) \leq \epsilon(h_1\,, h_3) + \epsilon(h_3\,, h_2)$. Then
\begin{small}
    \begin{align}
        \varsigma & \triangleq \mathop{min}\limits_{h \in \mathcal{H}} \left\{\epsilon_s(h, h_{\mathcal{S}}) + \epsilon_t(h, h_{\mathcal{T}})\right\} \nonumber \\
        & \leq \mathop{min}\limits_{h \in \mathcal{H}} \left\{\epsilon_s(h, h_{\mathcal{S}}) + \epsilon_t(h, h_{\mathcal{S}})  + \epsilon_t(h_{\mathcal{S}}, h_{\mathcal{T}}) \right\} \label{eq:three_triangle} \\
        & \leq \mathop{min}\limits_{h \in \mathcal{H}} \left\{\epsilon_s(h, h_{\mathcal{S}}) + \epsilon_t(h, h_{\mathcal{S}})  + \epsilon_t(h_{\mathcal{S}}, h_{\hat{\mathcal{T}}}) + \epsilon_t(h_{\hat{\mathcal{T}}}, h_{\mathcal{T}}) \right\} \,.
        \label{eq:triangle}
    \end{align}%
\end{small}%
In SDCA, the first and second terms in Eq.~\eqref{eq:triangle} should be small, since it is easily to find a $h$ to be agreed with $h_{\mathcal{S}}$ with accessing numerous labeled source data. Obviously the last term denotes the false pseudo rate
in our method which would be minimized using self-supervised learning strategy. Hence, the third term in Eq.~\eqref{eq:triangle}, which denotes the disagreement between labeling function $h_{\mathcal{S}}$ and $h_{\mathcal{T}}$ on target data, now should be payed more attention. Fortunately, the pixel-wise representation matching guided by semantic distributions enables driving the source and target pixel-wise representations towards corresponding semantic distribution. Consequently, $\epsilon_t(h_{\mathcal{S}}, h_{\hat{\mathcal{T}}})$ is expected to be small. 

As such, the SDCA constantly tightens the bound in Eq.~\eqref{eq-target-error-bound}, which provides theoretical insurance for SDCA to achieve expected segmentation performance.

%% file: 04-Experiment.tex
\section{Experiment}
\label{sec:experiment}
In this section, we compare the proposed method with several state-of-the-art domain adaptive semantic segmentation methods to demonstrate our advantages. We have performed comprehensive comparisons on multiple representative benchmarks, i.e., SYNTHIA $\to$ Cityscapes, GTAV $\to$ Cityscapes, and Cityscapes $\to$ Cross-City (Rio, Rome, Taipei and Tokyo). TABLE~\ref{table:datasets} shows the
statistics and configurations of the datasets. We first briefly describe datasets and evaluation metrics, and then provide implementation details. Next, the numeric experimental results of our model are presented for comparison with other state-of-the-art approaches. Finally, we further conduct analyses to understand the effect of each component in our method. The code is publicly available at \url{https://github.com/BIT-DA/SDCA}.

\subsection{Experimental Setups}
\subsubsection{Datasets}
{\bf Cityscapes~\cite{Cordts2016Cityscapes}} is a dataset of real urban scenes captured from 50 cities in Germany and neighboring countries. It consists of 2,975 training images, 500 validation images, and 1,525 test images, with the resolution at 2048 $\times$ 1024. All of these images are finely tagged with pixel-level semantic labels, and each pixel of the image is divided into 19 categories. For synthetic-to-real adaptation, since its test set does not provide ground truth labeling, we conduct evaluations on its validation set.  For cross city adaptation, following~\cite{tsai2018learning,wang2020class}, we use all images from the training set as the source training data.

{\bf SYNTHIA~\cite{ros2016synthia}} is a synthetic urban scene dataset. Following the prior works~\cite{tsai2018learning,luo2019taking}, we select its subset, called SYNTHIA-RAND-CITYSCAPES, that has 16 common semantic annotations with Cityscapes. In total, 9,400 images with the resolution 1280 $\times$ 760 from SYNTHIA dataset are used as the source domain data.

{\bf GTAV~\cite{stephan2016gtav}} is a composite image dataset sharing 19 semantic categories with Cityscapes. 24,966 urban scene images are extracted from the physically-based rendered computer game ``Grand Theft Auto V'' and are used as source domain data for training. 

{\bf Cross-City~\cite{chen2017nomorediscrimination}} is another real-world dataset collected from Google Street View. It provides four cities, i.e., Rio, Rome, Tokyo, and Taipei. For each city, there are 1,600 unlabeled image pairs which are taken at the same location but different times and all of them are used as target training data. Besides, for evaluation purpose~\cite{chen2017nomorediscrimination}, 100 images for each city are annotated with Cityscapes-compatible labeling.

\subsubsection{Evaluation Protocols}
The segmentation performance of our method in multiple scenarios are evaluated by the broadly utilized protocols, per-class intersection-over-union (IoU) and mean IoU over all categories. We use the evaluation code released along with the Cityscapes dataset, which calculates the PASCAL VOC intersection-over-union~\cite{everingham2015IoU}, i.e, IoU =$\frac{TP}{TP+FP+FN}$, where $TP$, $FP$, and $FN$ stand for the amount of true positive, false positive and false negative pixels, respectively. Before calculating the evaluation metric against the ground truth annotations, we upsample the segmentation outputs into the original resolution of input image.

\input{table/table_datasets.tex}

\input{table/table_synthetic.tex}

\subsubsection{Implementation Details}
{\bf Network architecture.} For fair comparison, we utilize the DeepLab-v2 framework~\cite{chen2018deeplab} with VGG-16~\cite{simonyan2015vgg} or ResNet-101~\cite{he2016deep} encoder as our segmentation network. All models are pre-trained on ImageNet~\cite{deng2009imagenet}. Subsequently, Atrous Spatial Pyramid Pooling (ASPP)~\cite{chen2018deeplab} is employed after the last layer of encoder with dilated rates \{6, 12, 18, 24\}. Lastly, an up-sampling layer along with a softmax operator is used to obtain the final pixel predictions with the input size.

{\bf Training details.} Our method is implemented with the PyTorch~\cite{paszke2019pytorch} library on Nvidia Tesla V100. To train the segmentation network $\Phi_\theta$, the Stochastic Gradient Descent is used as our optimizer, where the momentum is 0.9 and the weight decay is $10^{-4}$. The learning rate is initially set to $2.5\times 10^{-4}$ and we adopt polynomial learning rate scheduling with the power of 0.9. In all of our experiments, we set $\delta\,,\lambda_{lov}\,,\lambda_{feat}\,,\lambda_{seg}$ to 0.9, 0.75, 1.0, 1.0.
Regarding the training procedure, we first use the source data to train the network according to Eq.~\eqref{eq:ce_loss} as the baseline. Then the network is trained using our method for 40k iterations with batch size of 8 (four are source images and the other four are target images). Some strong data augmentations (e.g., color jittering and gaussian blur etc.) are applied for source images to prevent overfitting. Finally, we apply a simple self-supervised loss to further improve the performance. 

\subsection{Experimental Results}
\subsubsection{Results on SYNTHIA to Cityscapes}
First of all, we verify the effectiveness of out method in the SYNTHIA $\to$ Cityscapes task, and the comparison results are reported in TABLE~\ref{table:synthetic} with the best results highlighted in bold. We divide the TABLE~\ref{table:synthetic} into two parts according to different encoders (VGG-16 and ResNet-101). Overall, our SDCA framework surpasses all other approaches with a promising mIoU of 43.5\% (VGG-16), outperforming the baseline model trained merely on source data by a large increment of +19.0\% in mIoU. The improvement on another encoder ResNet-101 is similar. 

\begin{figure*}
    \centering
    \includegraphics[width=0.96\textwidth]{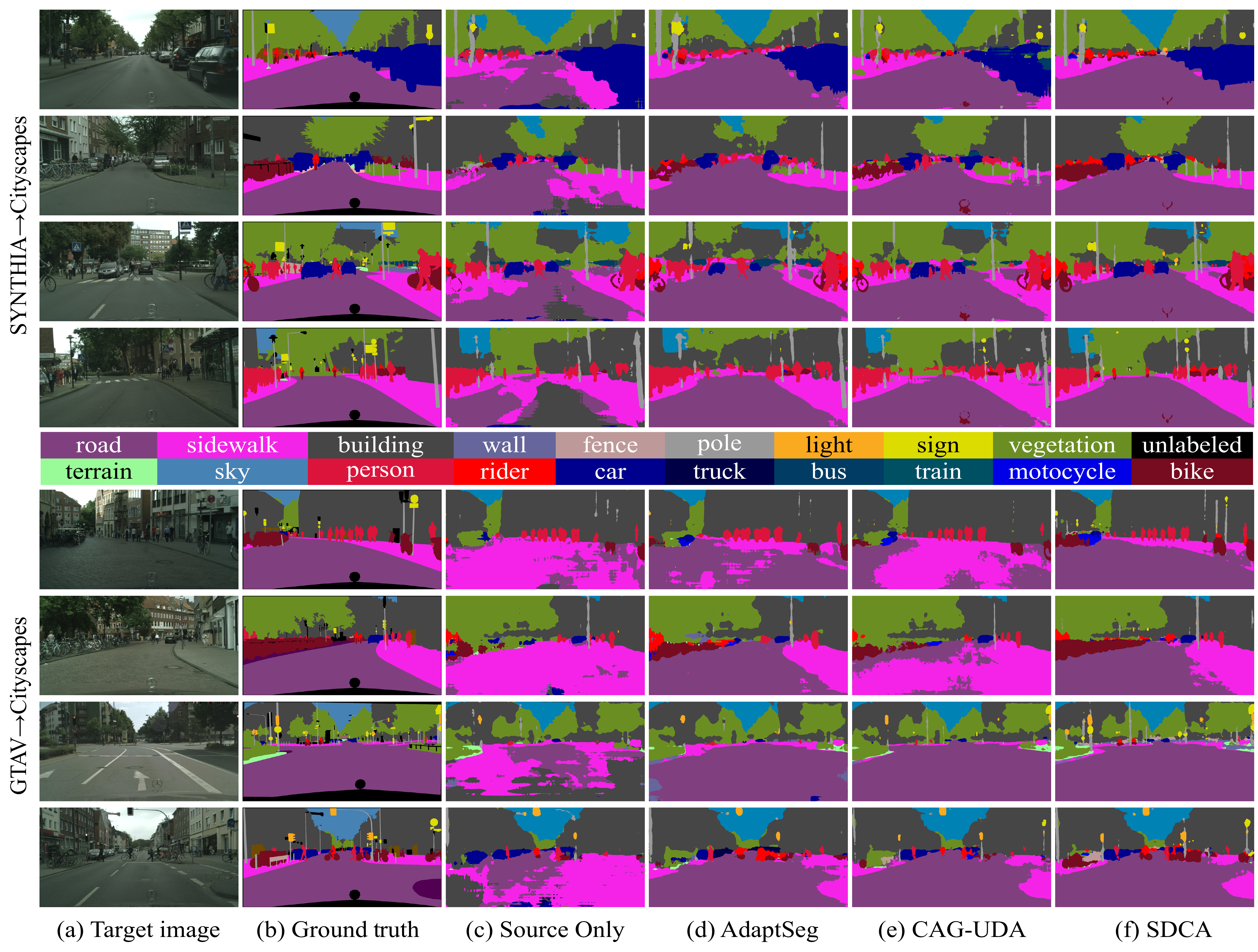}\vspace{-2mm}
    \caption{Qualitative results of semantic segmentation on the Cityscapes dataset. From the left to the right: target image, ground truth, the maps predicted by Source Only, AdaptSegNet~\cite{tsai2018learning}, CAG-UDA~\cite{zhang2019category} and our SDCA are shown one by one.}\vspace{-2mm}
    \label{Fig_seg_map}
\end{figure*}

It is interesting to observe that category alignment methods~\cite{kang2020pixel,luo2019taking,du2019ssf-dan,wang2020class} achieve better performance than global alignment counterparts~\cite{tsai2018learning,pan2020unsupervised,vu2019advent}, which proves that without considering the underlying structures among categories will inhibit the generalization performance. Relative to the self-supervised learning methods~\cite{zou2019confidence,lian2019pycda,li2019bidirectional}, the proposed method shows superior performance. For instance, PyCDA~\cite{lian2019pycda} leverages the pyramid curriculum to provide various properties about the target domain to guide the adaptation, yielding the mIoU scores 46.7\% and 53.3\% over the 16 and 13 categories (ResNet-101), which is inferior to our approach. Our SDCA model, as a category alignment method, also exceeds other related works~\cite{zhang2019category,wang2020differential,kang2020pixel}, demonstrating the effectiveness of the proposed method in learning discriminative pixel representations. We further show some of the segmentation samples in Fig.~\ref{Fig_seg_map} to qualitatively testify the superiority of our method.

\begin{figure}[!htbp]
    \centering
    \includegraphics[width=0.48\textwidth]{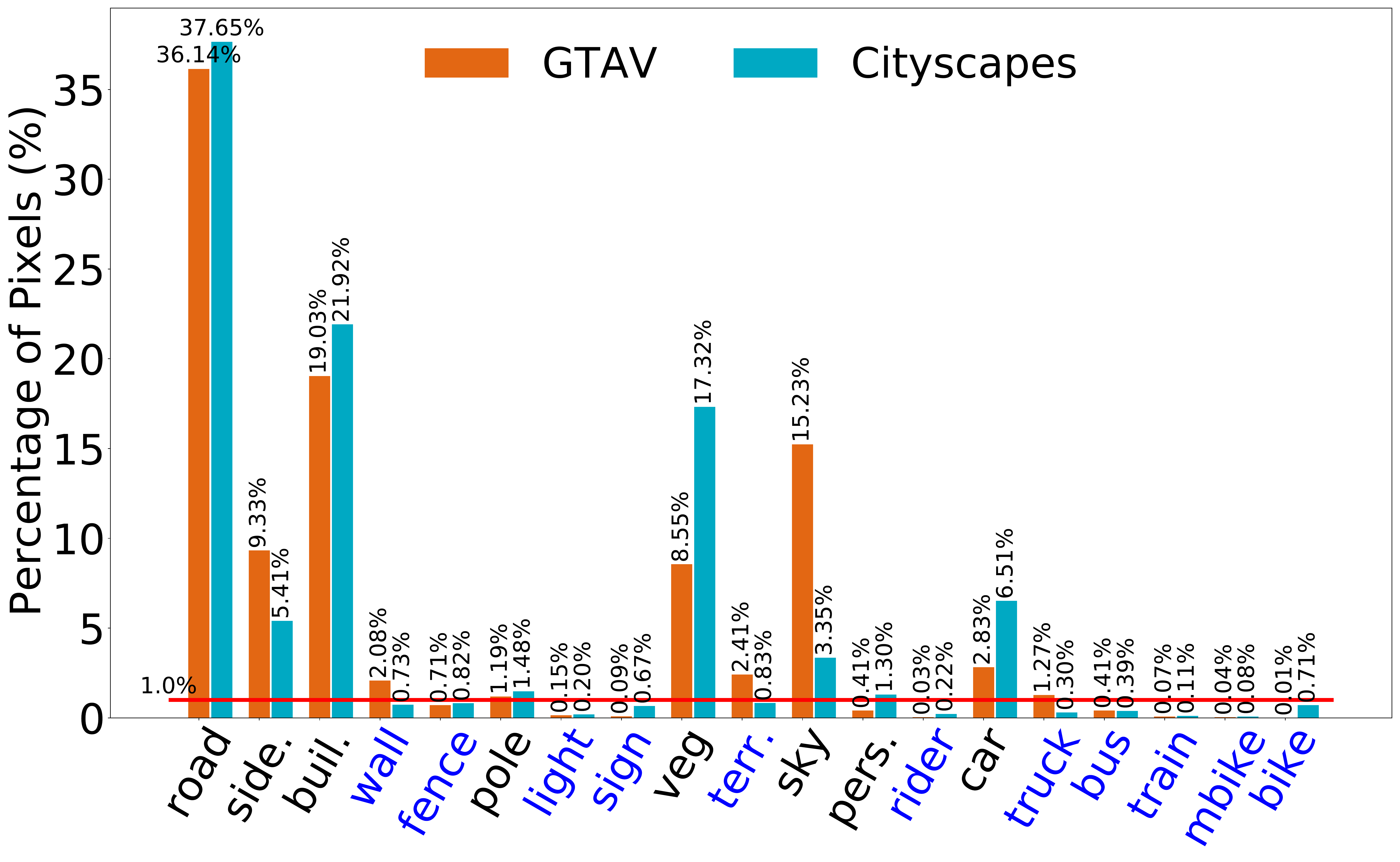}\vspace{-2mm}
    \caption{Class distributions on GTAV and Cityscapes datasets. In this paper, classes with a percentage of pixels less than 1.0\% are defined as the tail classes highlighted in \textcolor[rgb]{0, 0, 1}{blue}.}
    \label{Fig_tail_class}\vspace{-2mm}
\end{figure}
\input{table/table_gta.tex}
\subsubsection{Results on GTAV to Cityscapes}
TABLE~\ref{table:gta} demonstrates the semantic segmentation performance on GTAV $\to$ Cityscapes task in comparison with existing state-of-the-art domain adaptation approaches. We could observe that our SDCA achieves a significant performance gain, which is +3.5\% higher than FADA~\cite{wang2020class} for VGG-16 and is +2.7\% higher than CAG-UDA~\cite{zhang2019category} for ResNet-101 in terms of mIoU. In particular, our method also delivers evident advantages in the tail classes, which are highlighted in blue (refer to Fig.~\ref{Fig_tail_class} for more details). Specifically, despite the class ``bike" accounts for only 0.01\% ratio in GTAV class distribution, our SDCA can surpasses Source Only model with increments of 34.3\% and 37.0\% for VGG-16 and ResNet-101 respectively. In addition, for small objects such as poles and traffic signs, pervious methods may not adapt well as they easily get merged with background categories. In essence, the performance improvement of SDCA mostly comes from these challenging cases, because the proposed distribution-aware contrastive adaptation regards different categories equally. We also give the qualitative results of semantic segmentation on the Cityscapes validate set in Fig.~\ref{Fig_seg_map}, indicating that SDCA can preferably align the two domains in semantic level.

\input{table/table_cross_city.tex}
\subsubsection{Results on Cityscapes to Cross-City}
Except for the ``Sim-to-Real" scenarios for a large domain shift, we conduct experiment for small shift between real urban scenes from different cities (Rio, Rome, Taipei, and Tokyo)~\cite{chen2017nomorediscrimination}.
As shown in TABLE~\ref{tab:crosscity}, it is observed that our method exhibits superior segmentation performance in comparison to existing methods for different cities. On average over four cities, our SDCA achieves +6.4\% increment compared with the Source Only baseline, and +1.4\% gain compared with FADA~\cite{wang2020class}, the best baseline.

%% file: table/table_datasets.tex
\begin{table}
    \centering
    \caption{Statistics and configurations of the datasets.}\vspace{-2mm}
    \label{table:datasets}
    \resizebox{0.48\textwidth}{!}{
        \begin{tabular}{l c c c c}
            \toprule[1.0pt]
            Dataset & Cityscapes & GTAV & SYNTHIA & Cross-City\\
            \midrule
            \#Sample & 6,000 & 24,966 & 9,400 & 13,200 \\
            \#Class & 19 & 19 & 16 & 13 \\
            Resolution & 2048$\times$1024 & 1914$\times$1052 & 1280$\times$760 & 1295$\times$647 \\
            Training size & 1024$\times$512 & 1280$\times$720 & 1280$\times$760 & 1024$\times$512 \\
            Test size & 1024$\times$512 & - & - & 1024$\times$512 \\
            \bottomrule[1.0pt]
        \end{tabular}
    }
\end{table}

%% file: table/table_synthetic.tex
\begin{table*}
    \begin{center}
        \caption{Comparison results of {\bf SYNTHIA$\to$Cityscapes}. mIoU$^{*}$ denotes the mean IoU of 13 classes, excluding the classes with $^{*}$. ``GA", ``CA" and ``SSL" denote global alignment, category alignment and self-supervised learning respectively. 
        } \label{table:synthetic}
        \vspace{-2mm}
        \resizebox{\textwidth}{!}{
            \begin{tabular}{c l |c | c c c c c c c c c c c c c c c c | c c}
                \toprule[1.2pt]
                & Method & \rotatebox{60}{Mech.} & \rotatebox{60}{road} & \rotatebox{60}{side.} & \rotatebox{60}{buil.} & \rotatebox{60}{wall$^{*}$} & \rotatebox{60}{fence$^{*}$} & \rotatebox{60}{pole$^{*}$} & \rotatebox{60}{light} & \rotatebox{60}{sign} & \rotatebox{60}{veg.} & \rotatebox{60}{sky} & \rotatebox{60}{pers.} & \rotatebox{60}{rider} & \rotatebox{60}{car}&  \rotatebox{60}{bus} & \rotatebox{60}{mbike} & \rotatebox{60}{bike} & mIoU & mIoU$^{*}$ \\
                \midrule
                \multirow{11}{*}{\rotatebox{90}{VGG-16}} 
                & FCNs ITW~\cite{hoffman2016fcns} & GA & 11.5 & 19.6 & 30.8 & 4.4 & 0.0 & 20.3 & 0.1 & 11.7 & 42.3 & 68.7 & 51.2 & 3.8 & 54.0 & 3.2 & 0.2 & 0.6 & 20.2 & 22.9 \\
                & AdvEnt~\cite{vu2019advent} & GA & 67.9 & 29.4 & 71.9 & 6.3& 0.3& 19.9& 0.6& 2.6& 74.9& 74.9& 35.4& 9.6& 67.8& 21.4& 4.1& 15.5& 31.4& 36.6 \\   
                & AdaptSegNet~\cite{tsai2018learning} & GA & 78.9 & 29.2 & 75.5 & - & - & - & 0.1 & 4.8 & 72.6 & 76.7 & 43.4 & 8.8 & 71.1 & 16.0 & 3.6 & 8.4 &- & 37.6 \\
                \cmidrule(lr){2-21}  
                & Curr. DA~\cite{zhang2020curriculum} & SSL & 57.4 & 23.1 & 74.7 & 0.5 & 0.6 & 14.0 & 5.3 & 4.3 & 77.8 & 73.7 & 45.0 & 11.0 & 44.8 & 21.2 & 1.9 & 20.3 & 29.7 &  35.4\\    
                & CBST~\cite{zou2018unsupervised} & SSL & 69.6 & 28.7 & 69.5 & \bf 12.1 & 0.1 & 25.4 & 11.9 & 13.6 & 82.0 & 81.9 & 49.1 & 14.5 & 66.0 & 6.6 & 3.7 & 32.4 & 35.4 & 36.1 \\
                & BDL~\cite{li2019bidirectional} & SSL & 72.0 & 30.3 & 74.5 & 0.1 & 0.3 & 24.6 & 10.2 & \bf 25.2 & 80.5 & 80.0 & 54.7 & \bf 23.2 & 72.7 & 24.0 & 7.5 & 44.9 & 39.0 & 46.1\\
                \cmidrule(lr){2-21}
                & CLAN~\cite{luo2019taking} & CA & 80.4 & 30.7 &74.7&-&-&-& 1.4& 8.0& 77.1& 79.0& 46.5& 8.9& 73.8& 18.2& 2.2& 9.9&-& 39.3\\ 
                & SSF-DAN~\cite{du2019ssf-dan} & CA & \bf 87.1 & 36.5 & 79.7 & - & - & - & 13.5 & 7.8 & 81.2 & 76.7 & 50.1 & 12.7 & 78.0 & \bf 35.0 & 4.6 & 1.6 & - & 43.4 \\  
                & FADA~\cite{wang2020class} & CA & 80.4 & 35.9 & 80.9 & 2.5 & 0.3 &  30.4 & 7.9 & 22.3 & 81.8 & \bf 83.6 & 48.9 & 16.8 & 77.7 & 31.1 & \bf 13.5 & 17.9 & 39.5 & 46.0 \\ 
                \cmidrule(lr){2-21}
                & Source Only & - & 11.5 & 16.5 & 48.7 & 4.5 & 0.0 & 15.3 & 0.0 & 4.9 & 72.5 & 76.9 & 44.0 & 8.1 & 64.4 & 16.9 & 1.8 & 5.4 & 24.5 & 28.6\\     
                & Ours & CA & 86.2 & \bf 41.9 & \bf 82.1 & 5.3 & \bf 2.9 & \bf 34.1 & \bf 13.9 & 15.5 & \bf 82.8 & 82.0 & \bf 60.7 & 19.6 & \bf 81.4 & 33.3 & 8.6 & \bf 45.5 & \bf 43.5 & \bf 50.3 \\
                \midrule
                \midrule
                \multirow{15}{*}{\rotatebox{90}{ResNet-101}} 
                & AdaptSegNet~\cite{tsai2018learning} & GA & 79.2 & 37.2 & 78.8 & 10.5 & 0.3 & 25.1 & 9.9 & 10.5 & 78.2 & 80.5 & 53.5 & 19.6 & 67.0 & 29.5 & 21.6 & 31.3 & 39.5 & 45.9 \\ 
                & DS~\cite{dundar2020stylization} & GA & \bf 89.4 & \bf 51.9 & 80.0 & - & - & - & 2.3 & 5.1 & 79.4 & 83.1 & 48.9 & 18.3 & 79.3 & 36.9 & 19.4 &23.4 & - & 47.5 \\
                & AdvEnt~\cite{vu2019advent} & GA & 87.0 & 44.1 & 79.7 & 9.6 & 0.6 & 24.3 & 4.8 & 7.2 & 80.1 & 83.6 & 56.4 & 23.7 & 72.7 & 32.6 & 12.8 & 33.7 & 40.8 & 47.6 \\
                & IntraDA~\cite{pan2020unsupervised} & GA & 84.3 & 37.7 & 79.5 & 5.3 & 0.4 & 24.9 & 9.2 & 8.4 & 80.0 & 84.1 & 57.2 & 23.0 & 78.0 & 38.1 & 20.3 & 36.5 & 41.7 & 48.9 \\
                \cmidrule(lr){2-21}
                & CBST~\cite{zou2018unsupervised} & SSL & 68.0 & 29.9 & 76.3 & 10.8 & 1.4 & 33.9 & 22.8 & 29.5 & 77.6 & 78.3 & 60.6 & 28.3 & 81.6 & 23.5 & 18.8 & 39.8 & 42.6 & 48.9 \\
                & CRST~\cite{zou2019confidence} & SSL & 67.7 & 32.2 & 73.9 & 10.7 & 1.6 & 37.4 & 22.2 & 31.2 & 80.8 & 80.5 & 60.8 & 29.1 & 82.8 & 25.0 & 19.4 & 45.3 & 43.8 & 50.1 \\
                & BDL~\cite{li2019bidirectional} & SSL & 86.0 & 46.7 & 80.3 & - & - & - & 14.1 & 11.6 & 79.2 & 81.3 & 54.1 & 27.9 & 73.7 & 42.2 & 25.7 & 45.3 & - & 51.4 \\
                & PyCDA~\cite{lian2019pycda} & SSL & 75.5 & 30.9 & 83.3 & 20.8 & 0.7 & 32.7 & \bf 27.3 & \bf 33.5 & 84.7 & 85.0 & 64.1 & 25.4 & 85.0 & 45.2 & 21.2 & 32.0 & 46.7 & 53.3 \\
                \cmidrule(lr){2-21}
                & CLAN~\cite{luo2019taking} & CA & 81.3 & 37.0 & 80.1 & - &- & - & 16.1 &13.7 &78.2 & 81.5 & 53.4 & 21.2 & 73.0 & 32.9 & 22.6 & 30.7 & - & 47.8 \\ 
                & SSF-DAN~\cite{du2019ssf-dan} & CA & 84.6 & 41.7 & 80.8 & - &- & - & 11.5 & 14.7 & 80.8 & \bf 85.3 & 57.5 & 21.6 & 82.0 & 36.0 & 19.3 & 34.5 & - & 50.0 \\
                & SIM~\cite{wang2020differential} & CA & 83.0 & 44.0 & 80.3 & - &- & - & 17.1 & 15.8 & 80.5 & 81.8 & 59.9 & \textbf{33.1} & 70.2 & 37.3 & 28.5 & 45.8 & - & 52.1 \\
                & FADA~\cite{wang2020class} & CA & 84.5 & 40.1 & 83.1 & 4.8 & 0.0 & 34.3 & 20.1 & 27.2 & 84.8 & 84.0 & 53.5 & 22.6 & 85.4 & 43.7 & 26.8 & 27.8 & 45.2 & 52.5 \\ 
                & CAG-UDA~\cite{zhang2019category} & CA & 84.8 & 41.7 & \bf 85.5 & - & - & - & 13.7 & 23.0 & \bf 86.5 & 78.1 & 66.3 & 28.1 & 81.8 & 21.8 & 22.9 & 49.0 & - & 52.6 \\
                & PLCA~\cite{kang2020pixel} & CA & 82.6 & 29.0 & 81.0 & 11.2 & 0.2 & 33.6 & 24.9 & 18.3 & 82.8 & 82.3 & 62.1 & 26.5 & 85.6 & \bf 48.9 & 26.8 & 52.2 & 46.8 & 54.0 \\
                \cmidrule(lr){2-21}
                & Source Only & - & 40.9 & 21.6 & 60.6 & 10.2 & 0.0 & 26.0 & 5.9 & 14.4 & 77.9 & 69.8 & 52.8 & 15.8 & 71.9 & 35.0 & 8.5 & 25.3 & 33.5 & 38.5 \\ 
                & Ours & CA & 88.4 & 45.9 & 83.9 & \bf 24.0 &  \bf 1.7 & \bf 38.1 & 25.2 &  17.0 &  85.3 &  82.9 &  \bf 67.3 &  26.6 & \bf 87.1 & 47.2 & \bf 28.6 & \bf 53.4 &  \bf 50.2 & \bf 56.8 \\         
                \bottomrule[1.2pt]
                \end{tabular}
        }
    \end{center}
    \vspace{-3mm}
\end{table*}

%% file: table/table_gta.tex
\begin{table*}
    \begin{center}
        \caption{Comparison results of {\bf GTAV$\to$Cityscapes}. mIoU$^{t}$ denotes the mean IoU of the tail classes in \textcolor{blue}{blue}.} \label{table:gta} \vspace{-2mm}
        \resizebox{\textwidth}{!}{
        \begin{tabular}{c l |c| c c c c c c c c c c c c c c c c c c c |c c}
            \toprule[1.2pt]
            & Method & \rotatebox{60}{Mech.} & \rotatebox{60}{road} & \rotatebox{60}{side.} & \rotatebox{60}{buil.} & \rotatebox{60}{\textcolor{blue}{wall}} & \rotatebox{60}{\textcolor{blue}{fence}} & \rotatebox{60}{pole} & \rotatebox{60}{\textcolor{blue}{light}} & \rotatebox{60}{\textcolor{blue}{sign}} & \rotatebox{60}{veg.} & \rotatebox{60}{\textcolor{blue}{terr.}} & \rotatebox{60}{sky} & \rotatebox{60}{pers.} & \rotatebox{60}{\textcolor{blue}{rider}} & \rotatebox{60}{car}& \rotatebox{60}{\textcolor{blue}{truck}} & \rotatebox{60}{\textcolor{blue}{bus}} & \rotatebox{60}{\textcolor{blue}{train}} & \rotatebox{60}{\textcolor{blue}{mbike}} & \rotatebox{60}{\textcolor{blue}{bike}} & mIoU & mIoU$^{t}$ \\
            \midrule
            \multirow{11}{*}{\rotatebox{90}{VGG-16}} 
            & FCNs ITW~\cite{hoffman2016fcns} & GA & 70.4 & 32.4 & 62.1 & 14.9 & 5.4 & 10.9 & 14.2 & 2.7 & 79.2 & 21.3 & 64.6 & 44.1 & 4.2 & 70.4 & 8.0 & 7.3 & 0.0 & 3.5 & 0.0 & 27.1 & 7.4 \\
            & AdaptSegNet~\cite{tsai2018learning} & GA & 87.3 & 29.8 & 78.6 & 21.1 & 18.2 & 22.5 & 21.5 & 11.0 & 79.7 & 29.6 & 71.3 & 46.8 & 6.5 & 80.1 & 23.0 & 26.9 & 0.0 & 10.6 & 0.3 & 35.0 & 15.3 \\
            & AdvEnt~\cite{vu2019advent} & GA & 86.9 & 28.7 & 78.7 & 28.5 & 25.2 & 17.1 & 20.3 & 10.9 & 80.0 & 26.4 & 70.2 & 47.1 & 8.4 & 81.5 & 26.0 & 17.2 & \bf 18.9 & 11.7 & 1.6 & 36.1 & 17.7 \\
            \cmidrule(lr){2-24} 
            & Curr. DA~\cite{zhang2020curriculum} & SSL & 72.9 & 30.0 & 74.9 & 12.1 & 13.2 & 15.3 & 16.8 & 14.1 & 79.3 & 14.5 & 75.5 & 35.7 & 10.0 & 62.1 & 20.6 & 19.0 & 0.0 & 19.3 & 12.0 & 31.4 & 13.8 \\ 
            & CBST~\cite{zou2018unsupervised} & SSL & 90.4 & 50.8 & 72.0 & 18.3 & 9.5 & 27.2 & 28.6 & 14.1 & 82.4 & 25.1 & 70.8 & 42.6 & 14.5 & 76.9 & 5.9 & 12.5 & 1.2 & 14.0 & 28.6 & 36.1 & 15.7 \\  
            & BDL~\cite{li2019bidirectional} & SSL & 89.2 & 40.9 & 81.2 & 29.1 & 19.2 & 14.2 & 29.0 & 19.6 & \bf 83.7 & 35.9 & 80.7 & 54.7 & 23.3 & 82.7 & 25.8 & 28.0 & 2.3 & \bf 25.7 & 19.9 & 41.3 & 23.4\\
            \cmidrule(lr){2-24} 
            & CLAN~\cite{luo2019taking} & CA & 88.0 & 30.6 & 79.2 & 23.4 & 20.5 & 26.1 & 23.0 & 14.8 & 81.6 & 34.5 & 72.0 & 45.8 & 7.9 & 80.5 & 26.6 & 29.9 & 0.0 & 10.7 & 0.0 & 36.6 & 17.4 \\
            & SSF-DAN~\cite{du2019ssf-dan} & CA & 88.7 & 32.1 & 79.5 & 29.9 & 22.0 & 23.8 & 21.7 & 10.7 & 80.8 & 29.8 & 72.5 & 49.5 & 16.1 & 82.1 & 23.2 & 18.1 & 3.5 & 24.4 & 8.1 & 37.7 & 18.9\\ 
            & SIM~\cite{wang2020differential} & CA & 88.1 & 35.8 & 83.1 & 25.8 & 23.9 & 29.2 & 28.8 & 28.6 & 83.0 & \bf 36.7 & 82.3 & 53.7 & 22.8 & 82.3 & 26.4 & 38.6 & 0.0 & 19.6 & 17.1 & 42.4 & 24.4\\ 
            & FADA~\cite{wang2020class} & CA & 92.3 & 51.1 & \bf 83.7 & \bf 33.1 & \bf 29.1 & 28.5 & 28.0 & 21.0 & 82.6 & 32.6 & \bf 85.3 & 55.2 & 28.8 & 83.5 & 24.4 & 37.4 & 0.0 & 21.1 & 15.2 & 43.8 & 24.6 \\
            \cmidrule(lr){2-24} 
            & Source Only & - & 50.5 & 19.0 & 65.7 & 16.5 & 8.8 & 20.4 & 22.5 & 11.2 & 78.7 & 6.4 & 67.0 & 45.2 & 17.6 & 73.9 & 16.0 & 21.0 & 0.0 & 12.3 & 4.8 & 29.3 & 12.5\\
            & Ours & CA & \bf 93.2 & \bf 56.3 & 83.4 & 25.4 & 22.3 & \bf 36.7 & \bf 37.6 & \bf 32.2 &  83.3 & 33.3 & 83.2 & \bf 62.9 & \bf 34.6 & \bf 85.4 & 23.0 & \bf 41.6 & 0.0 & 24.2 & \bf 39.1 & \bf 47.3 & \bf 28.5 \\
            \midrule
            \midrule
            \multirow{15}{*}{\rotatebox{90}{ResNet-101}} 
            & AdaptSegNet~\cite{tsai2018learning} & GA & 86.5 & 36.0 & 79.9 & 23.4 & 23.3 & 23.9 & 35.2 & 14.8 & 83.4 & 33.3 & 75.6 & 58.5 & 27.6 & 73.7 & 32.5 & 35.4 & 3.9 & 30.1 & 28.1 & 42.4 & 26.1 \\
            & AdvEnt~\cite{vu2019advent}  & GA & 89.9 & 36.5 & 81.6 & 29.2 & 25.2 & 28.5 & 32.3 & 22.4 & 83.9 & 34.0 & 77.1 & 57.4 & 27.9 & 83.7 & 29.4 & 39.1 & 1.5 & 28.4 & 23.3 & 43.8 & 26.6 \\
            & DS~\cite{dundar2020stylization}  & GA & 91.5 & 51.8 & 83.0 & 35.8 & 21.5 & 31.7 & 35.0 & 22.2 & 83.8 & 39.4 & 82.1 & 59.5 & 29.3 & 82.7 & 28.6 & 35.2 & 2.9 & 22.5 & 17.5 & 45.1 & 26.4 \\ 
            & IntraDA~\cite{pan2020unsupervised}  & GA & 90.6 & 37.1 & 82.6 & 30.1 & 19.1 & 29.5 & 32.4 & 20.6 & 85.7 & 40.5 & 79.7 & 58.7 & 31.1 & 86.3 & 31.5 & 48.3 & 0.0 & 30.2 & 35.8 & 46.3 & 29.1 \\ 
            \cmidrule(lr){2-24} 
            & CBST~\cite{zou2018unsupervised} & SSL & 91.8 & 53.5 & 80.5 & 32.7 & 21.0 & 34.0 & 28.9 & 20.4 & 83.9 & 34.2 & 80.9 & 53.1 & 24.0 & 82.7 & 30.3 & 35.9 & 16.0 & 25.9 & 42.8 & 45.9 & 28.4 \\
            & CRST~\cite{zou2019confidence} & SSL & 91.0 & \bf 55.4 & 80.0 & 33.7 & 21.4 & 37.3 & 32.9 & 24.5 & 85.0 & 34.1 & 80.8 & 57.7 & 24.6 & 84.1 & 27.8 & 30.1 & 26.9 & 26.0 & 42.3 & 47.1 & 29.5 \\
            & PyCDA~\cite{lian2019pycda} & SSL & 90.5 & 36.3 & 84.4 & 32.4 & 28.7 & 34.6 & 36.4 & 31.5 & \bf 86.8 & 37.9 & 78.5 & 62.3 & 21.5 & 85.6 & 27.9 & 34.8 & 18.0 & 22.9 & 49.3 & 47.4 & 31.0 \\
            & BDL~\cite{li2019bidirectional} & SSL & 91.0 & 44.7 & 84.2 & 34.6 & 27.6 & 30.2 & 36.0 & 36.0 & 85.0 & \bf 43.6 & 83.0 & 58.6 & 31.6 & 83.3 & 35.3 & 49.7 & 3.3 & 28.8 & 35.6 & 48.5 & 32.9 \\
            \cmidrule(lr){2-24} 
            & CLAN~\cite{luo2019taking} & CA & 87.0 & 27.1 & 79.6 & 27.3 & 23.3 &28.3 & 35.5 & 24.2 & 83.6 & 27.4 & 74.2 & 58.6 & 28.0 & 76.2 &   33.1 & 36.7 & 6.7 & 31.9 & 31.4 & 43.2 & 27.8 \\
            & SSF-DAN~\cite{du2019ssf-dan} & CA & 90.3 & 38.9 & 81.7 & 24.8 & 22.9 & 30.5 & 37.0 & 21.2 & 84.8 & 38.8 & 76.9 & 58.8 & 30.7 & 85.7 & 30.6 & 38.1 & 5.9 & 28.3 & 36.9 & 45.4 & 28.7 \\           
            & PLCA~\cite{kang2020pixel} & CA & 84.0 & 30.4 & 82.4 & 35.3 & 24.8 & 32.2 & 36.8 & 24.5 & 85.5 & 37.2 & 78.6 & 66.9 & 32.8 & 85.5 & 40.4 & 48.0 & 8.8 & 29.8 & 41.8 & 47.7 & 32.7\\
            & SIM~\cite{wang2020differential} & CA & 90.6 & 44.7 & 84.8 & 34.3 & 28.7 & 31.6 & 35.0 & 37.6 & 84.7 & 43.3 & 85.3 & 57.0 & 31.5 & 83.8 & \bf 42.6 & 48.5 & 1.9 & 30.4 & 39.0 & 49.2 & 33.9 \\
            & FADA~\cite{wang2020class} & CA & 91.0 & 50.6 & \textbf{86.0} & \textbf{43.4} & \bf 29.8 & 36.8 & 43.4 & 25.0 & \textbf{86.8} & 38.3 & \textbf{87.4} & 64.0 & \textbf{38.0} & 85.2 & 31.6 & 46.1 & 6.5 & 25.4 & 37.1 & 50.1 & 33.1 \\  
            & CAG-UDA~\cite{zhang2019category} & CA & 90.4 & 51.6 & 83.8 & 34.2 & 27.8 & 38.4 & 25.3 & \bf 48.4 & 85.4 & 38.2 & 78.1 & 58.6 & 34.6 & 84.7 & 21.9 & 42.7 & \bf 41.1 & 29.3 & 37.2 & 50.2 & 34.6 \\
            \cmidrule(lr){2-24} 
            & Source Only & - & 70.2 & 14.6 & 71.3 & 24.1 & 15.3 & 25.5 & 32.1 & 13.5 & 82.9 & 25.1 & 78.0 & 56.2 & 33.3 & 76.3 & 26.6 & 29.8 & 12.3 & 28.5 & 18.0 & 38.6 &  23.5\\ 
            & Ours & CA & \bf 92.8 & 52.5 & 85.9 & 34.8 & 28.1 & \bf 40.3 & \bf 44.4 & 33.4 & 86.7 & 41.7 & 87.1 & \bf 67.4 & 37.7 & \bf 88.1 & 39.9 & \bf 52.5 & 1.4 & \bf 34.2 & \bf 55.0 & \bf 52.9 & \bf 36.7 \\
            \bottomrule[1.2pt]
        \end{tabular}
        }
    \end{center}
    \vspace{-2mm}
    \end{table*}

%% file: table/table_cross_city.tex
\begin{table*}
    \begin{center}
    \caption{Comparison results of {\bf Cityscapes $\to$ Cross-City}. We apply the DeepLab-V2 architecture as Source Only baseline and compare the results among state-of-the-art approaches.}
    \vspace{-2mm}
    \label{tab:crosscity}
    \resizebox{\textwidth}{!}{
    \renewcommand{\arraystretch}{1.2}
    \begin{tabular}{c l | c c c c c c c c c c c c c | c c}
    \toprule[1.2pt]
    City & Method & \rotatebox{60}{road} & \rotatebox{60}{side.} & \rotatebox{60}{buil.} & \rotatebox{60}{light} & \rotatebox{60}{sign} & \rotatebox{60}{veg} & \rotatebox{60}{sky} & \rotatebox{60}{pers.} & \rotatebox{60}{rider} & \rotatebox{60}{car} & \rotatebox{60}{bus} & \rotatebox{60}{mbike} & \rotatebox{60}{bike} & mIoU & gain\\
    \hline
    \multirow{7}{*}{Rio} & Dilated-FCN~\cite{yu2016dilated} & 69.0 & 31.8 & 77.0 & 4.7 & 3.7 & 71.8 & 80.8 & 38.2 & 8.0 & 61.2 & 38.9 & 11.5 & 3.4 & 38.5 & - \\
    & Cross-City \cite{chen2017nomorediscrimination} & 74.2 & 43.9 & 79.0 & 2.4 & 7.5 & 77.8 & 69.5 & 39.3 & 10.3 & 67.9 & \bf 41.2 & 27.9 & 10.9 & 42.5 & 4.0 \\
    \cmidrule(lr){2-17}
    & DeepLab-v2~\cite{chen2018deeplab} & 76.6 & 47.3 & 82.5 & 12.6 & 22.5 & 77.9 & 86.5 & 43.0 & 19.8 & 74.5 & 36.8 & 29.4 & 16.7 & 48.2 & - \\
    & AdaptSegNet \cite{tsai2018learning}& 76.2 & 44.7 & \bf 84.6 & 9.3 & 25.5 & \bf 81.8 & 87.3 & 55.3 & 32.7 & 74.3 & 28.9 & 43.0 & 27.6 & 51.6 & 3.4 \\
    & SSF-DAN~\cite{du2019ssf-dan} & 74.2 & 43.7 & 82.5 & 10.3 & 21.7 & 79.4 & 86.7 & 55.9 & \bf 36.1 & 74.9 & 33.7 & 52.6 & 33.7 & 52.7 & 4.5 \\
    & FADA~\cite{wang2020class} & 80.6 & 53.4 & 84.2 & 5.8 &23.0 & 78.4& 87.7 & 60.2 & 26.4 & 77.1 & 37.6 & \bf 53.7 & 42.3 & 54.7 & 6.5 \\	
    & Ours & \bf 81.5 & \bf 56.3 & 75.4 & \bf 14.8 & \bf 36.5 & 79.8 & \bf 89.2 & \bf 60.5 & 21.3 & \bf 80.4 & 39.1 & 44.9 & \bf 42.5 & \bf 55.5 & \bf 7.3 \\
    \hline
    \hline

    \multirow{7}{*}{Rome} 
    & Dilated-FCN~\cite{yu2016dilated} & 77.7& 21.9& 83.5& 0.1& 10.7& 78.9& 88.1& 21.6& 10.0& 67.2& 30.4& 6.1& 0.6& 38.2 & - \\
    & Cross-City \cite{chen2017nomorediscrimination} & 79.5 & 29.3 & 84.5 & 0.0 & 22.2 & 80.6 & 82.8 & 29.5 & 13.0 & 71.7 & 37.5 & 25.9 & 1.0 & 42.9 & \bf 4.5 \\
    \cmidrule(lr){2-17}
    & DeepLab-v2~\cite{chen2018deeplab} & 83.9& 34.3& 87.7& 13.0& 41.9 &84.6& 92.5& 37.7& 22.4 & 80.8 & 38.1 & 39.1 & 5.3 & 50.9 & - \\
    & AdaptSegNet \cite{tsai2018learning}& 83.9 & 34.2 &\textbf{88.3} & 18.8 & 40.2 & \textbf{86.2} & \textbf{93.1} & 47.8 & 21.7 & 80.9 & 47.8 & 48.3 & 8.6 & 53.8 & 2.9 \\
    & SSF-DAN~\cite{du2019ssf-dan} & 84.2 & 38.4 & 87.4 & \bf 23.4 & 43.0 & 85.6 & 88.2 & 50.2 & 23.7 & 80.6 & 38.1 & 51.6 & 8.6 & 54.1 & 3.2 \\
    & FADA~\cite{wang2020class} & 84.9 & 35.8 &\textbf{88.3} & 20.5 & 40.1 & 85.9 & 92.8 &\textbf{56.2} & 23.2 & \textbf{83.6} & 31.8 & \textbf{53.2} & \textbf{14.6} & 54.7 & 3.8 \\
    & Ours & \bf 87.8 & \bf 44.9 & 84.3 & 15.0 & \bf 43.7 & 85.3 & \bf 93.1 & 48.8 & \bf 28.2 & 81.8 & \bf 53.5 & 45.3 & 2.5 & \bf 54.9 & 4.0 \\
    \hline
    \hline

    \multirow{7}{*}{Taipei} 
    &Dilated-FCN~\cite{yu2016dilated} & 77.2& 20.9& 76.0& 5.9& 4.3& 60.3& 81.4& 10.9& 11.0& 54.9& 32.6& 15.3& 5.2& 35.1 & - \\
    & Cross-City \cite{chen2017nomorediscrimination} & 78.6& 28.6& 80.0& 13.1& 7.6& 68.2& 82.1& 16.8& 9.4& 60.4& 34.0& 26.5& 9.9& 39.6 & 4.5 \\
    \cmidrule(lr){2-17}
    & DeepLab-v2~\cite{chen2018deeplab} & 83.5 & 33.4 & 86.6 & 12.7 & 16.4 & 77.0 & 92.1 & 17.6 & 13.7 & 70.7 & 37.7 & 44.4 & 18.5 & 46.5 & - \\  
    & AdaptSegNet \cite{tsai2018learning} & 81.7& 29.5& 85.2& \textbf{26.4}& 15.6& 76.7& 91.7& 31.0& 12.5& 71.5& 41.1 & 47.3& 27.7& 49.1 & 2.6 \\
    & SSF-DAN~\cite{du2019ssf-dan} & 84.5 & 35.3 & \bf 86.4 & 17.7 & 16.9 & 77.7 & 91.3 & 31.8 & \bf 22.3 & \bf 73.7 & 41.1 & \bf 55.9 & 28.5 & 51.0 & 4.5 \\
    & FADA~\cite{wang2020class} &\textbf{86.0} & \textbf{42.3} & 86.1 & 6.2 & 20.5 & \textbf{78.3} & 92.7 & \textbf{47.2} & 17.7 & 72.2 & 37.2 & 54.3 & \textbf{44.0} & 52.7 & 6.2 \\
    & Ours & \bf 86.0 & 38.8 & 85.7 & 16.7 & \bf 24.0 & 76.4 & \bf 93.6 & 39.0 & 20.7 & 73.0 & \bf 52.3 & 55.2 & 35.3 & \bf 53.6 & \bf 7.1 \\
    \hline
    \hline
    
    \multirow{7}{*}{Tokyo} 
    & Dilated-FCN~\cite{yu2016dilated} & 81.2& 26.7& 71.7& 8.7 &5.6& 73.2& 75.7& 39.3& 14.9& 57.6& 19.0& 1.6& 33.8& 39.2 & - \\
    & Cross-City \cite{chen2017nomorediscrimination} & 83.4 & 35.4 & 72.8 & 12.3 & 12.7 & 77.4 & 64.3 & 42.7 & 21.5 & 64.1 & 20.8 & 8.9 & 40.3 & 42.8 & 3.6 \\
    \cmidrule(lr){2-17}
    & DeepLab-v2~\cite{chen2018deeplab} & 82.9 & 31.3 & 78.7 & 14.2 & 24.5 & 81.6 & 89.2 & 48.6 & 33.3 & 70.5 & 7.7 & 11.5 & 45.9 & 47.7 & - \\
    & AdaptSegNet \cite{tsai2018learning}& 81.5 & 26.0 & 77.8 & \bf 17.8 & \bf 26.8 & 82.7 & 90.9 & 55.8 & \bf 38.0 & 72.1 & 4.2 & 24.5 & 50.8 & 49.9 & 2.2 \\		
    & SSF-DAN~\cite{du2019ssf-dan} & 82.1 & 27.4 & 78.0 & 18.4 & 26.6 & 83.0 & 90.8 & 57.1 & 35.8 & 72.0 & 4.6 & 27.3 & 52.8 & 50.4 & 2.7 \\
    & FADA~\cite{wang2020class} & 85.8 & 39.5 & 76.0 & 14.7 & 24.9 & \textbf{84.6} & 91.7 & \textbf{62.2} & 27.7 & 71.4 & 3.0 & 29.3 & \textbf{56.3} & 51.3 & 3.6 \\	
    & Ours & \bf 87.8 & \bf 42.8 & \bf 86.4 & 12.5 & 26.1 & 76.7 & \bf 93.5 & 46.7 & 18.7 & \bf 74.4 & \bf 47.4 & \bf 58.3 & 44.1 & \bf 55.0 & \bf 7.3 \\
    \bottomrule[1.2pt]
    \end{tabular}
    }
    \end{center}
    \vspace{-2mm}
\end{table*}

%% file: 05-Analysis.tex
\subsection{Analysis}
\label{sec:analysis}

\subsubsection{Ablation Studies}
In this section, we perform extensive ablation experiments to inspect the role of each component present in the SDCA framework. Specifically, we testify SDCA on two ``Sim-to-Real'' scenarios, i.e., SYNTHIA $\to$ Cityscapes and GTAV $\to$ Cityscapes deployed on ResNet-101 and the results are presented in TABLE~\ref{table:ablation_synthetic} and TABLE~\ref{table:ablation_gta} respectively. 

\begin{figure*}[!htbp]
    \centering
     \includegraphics[width=\textwidth]{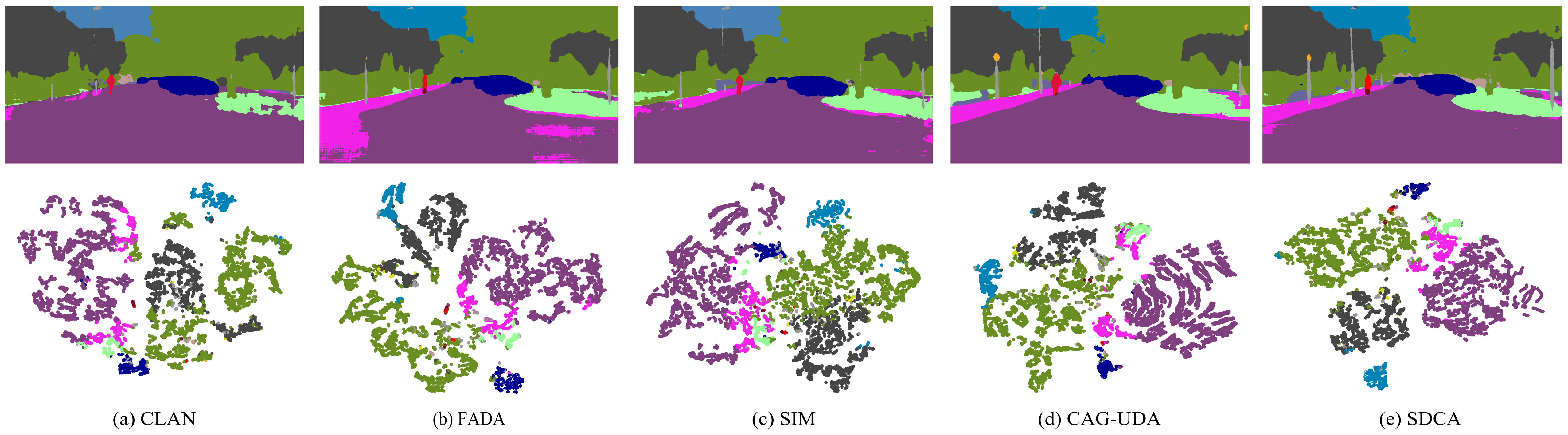}
    \vspace{-5mm}
    \caption{t-SNE analysis~\cite{maaten2008visualizing} of existing category alignment methods and SDCA. The visualization of the embedded features further proves that SDCA can exhibit the clearest clusters compared with other baseline methods.}
    \label{Fig_tsne}
\end{figure*}

\begin{figure*}
    \centering
    \includegraphics[width=\textwidth]{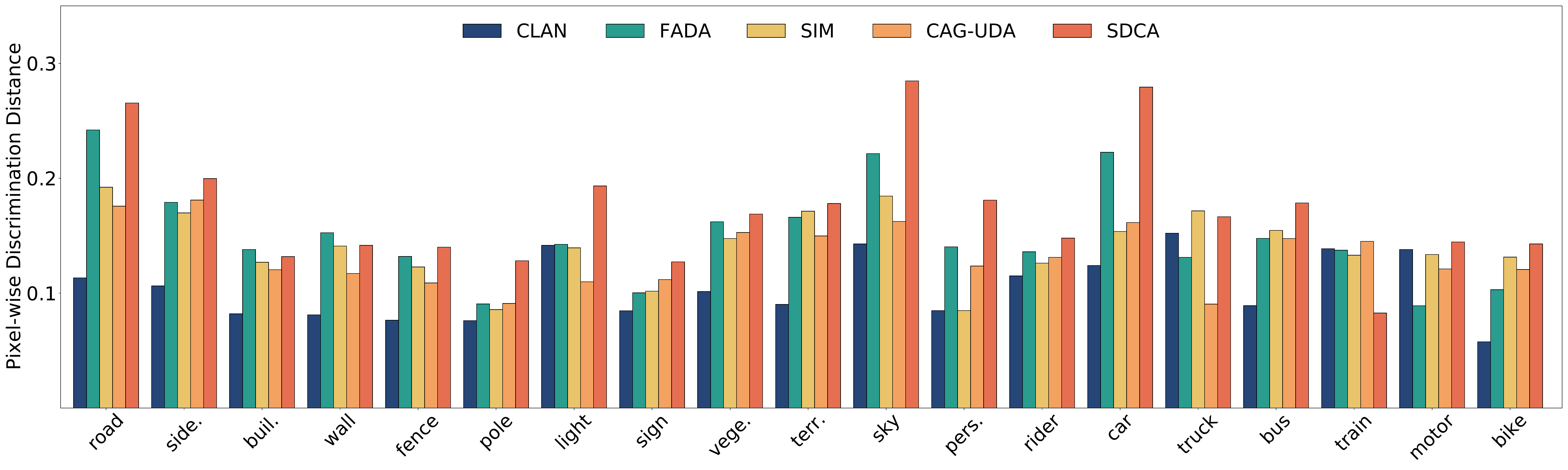}
    \vspace{-6mm}
    \caption{Quantitative analysis of the feature distributions. For each category, we show the values of pixel-wise discrimination distance (PDD) as defined in Eq.~\eqref{eq:pdd} on the whole Cityscapes validation set. These results are from 1) traditional category-level adversarial learning approaches, i.e, CLAN~\cite{luo2019taking}, FADA~\cite{wang2020class}, 2) category centroid based approaches i.e., CAG-UDA~\cite{zhang2019category}, SIM~\cite{wang2020differential} and 3) the adapted model using SDCA, respectively. A high PDD suggests the pixel-wise representations of same category are clustered densely while the distance between different categories is relatively large.}
    \label{Fig_ccd}
    \vspace{-2mm}
\end{figure*}

At first, we require a segmentation model purely trained on the source samples to investigate the primary performance gain that comes from the proposed framework, and we call this model the ``Source Only" model that gives the lower bound. TABLE~\ref{table:ablation_synthetic} and TABLE~\ref{table:ablation_gta} demonstrate the performance increments compared to the ``Source Only" baseline, by progressively adding each loss function into the pipeline. It can be seen that each module contributes to overall performance of our framework. Finally, we achieve 50.2\% mIoU and 56.8\% mIoU* with SYNTHIA as the source domain and 52.9\% mIoU and 36.7 mIoU$^{t}$ with GTA as the source domain, outperforming the ``Source Only" model by a significant margin. 

\input{table/table_ablation_study.tex}

{\bf Effect of contrastive adaptation.} As discussed in Section~\ref{sec:contrative_adaptation}, contrastive adaptation at pixel level guided by semantic distributions can build up stronger intra-/inter-category connections and minimize the domain divergence efficiently. In TABLE~\ref{table:ablation_synthetic} and TABLE~\ref{table:ablation_gta}, compared with ``Source Only + $\mathcal{L}_{lov}$", the model can achieve an improvement of +7.6\% mIoU for SYNTHIA $\to$ Cityscapes and an increment of +6.0\% mIoU for GTAV $\to$ Cityscapes, verifying the effectiveness of contrastive adaptation.

{\bf Effect of multi-level adaptation.} As shown on the second row from the bottom in TABLE~\ref{table:ablation_synthetic} and TABLE~\ref{table:ablation_gta}, additionally performing the contrastive adaptation on segmentation outputs, i.e., $O_s$ and $O_t$, gives rise to promising improvement, insinuating that the multi-level adaptations complement each other to yield discriminative pixel-wise representations and mitigate the domain shift.

{\bf Effect of self-supervised learning.} We respectively select pseudo labels according to ``Source Only" model and ``Ours (multi-level)" and then retrain the network with $\mathcal{L}_{ssl}$. In TABLE~\ref{table:ablation_synthetic} and TABLE~\ref{table:ablation_gta}, as expected, applying self-supervised loss on the target domain, the segmentation performance of the final models improve noticeably, implying the necessity of learning the intrinsic structure of the target domain. Moreover, our multi-level contrastive adaptation method equipped with a simple self-supervised learning strategy can further boost the performance.

\subsubsection{t-SNE Visualization}
To better develop intuition, we draw t-SNE visualizations~\cite{maaten2008visualizing} of the learned feature representations for related category alignment methods (CLAN~\cite{luo2019taking}, FADA~\cite{wang2020class}, SIM~\cite{wang2020differential}, CAG-UDA~\cite{zhang2019category}) and our SDCA in Fig.~\ref{Fig_tsne}. To this end, we first randomly select a image from target domain and then map its high-dimensional latent feature representations to a 2D space. From the t-SNE analysis, we can observe that previous category alignment methods could produce separated features, yet it may be hard for dense prediction since the margins between different category features are not obvious and the distribution is still dispersed. When we apply semantic distribution-aware contrastive adaptation, features among different categories are better separated, demonstrating that the semantic distributions can provide correct supervision signal for target data. In comparison, the embedded representations of our SDCA exhibit the clearest clusters compared with other state-of-the-arts, revealing the discriminative capability of the contrastive adaptation.

\begin{figure*}[!htbp]
    \centering
    \subfigure[$\delta$]{
        \includegraphics[width=0.23\textwidth]{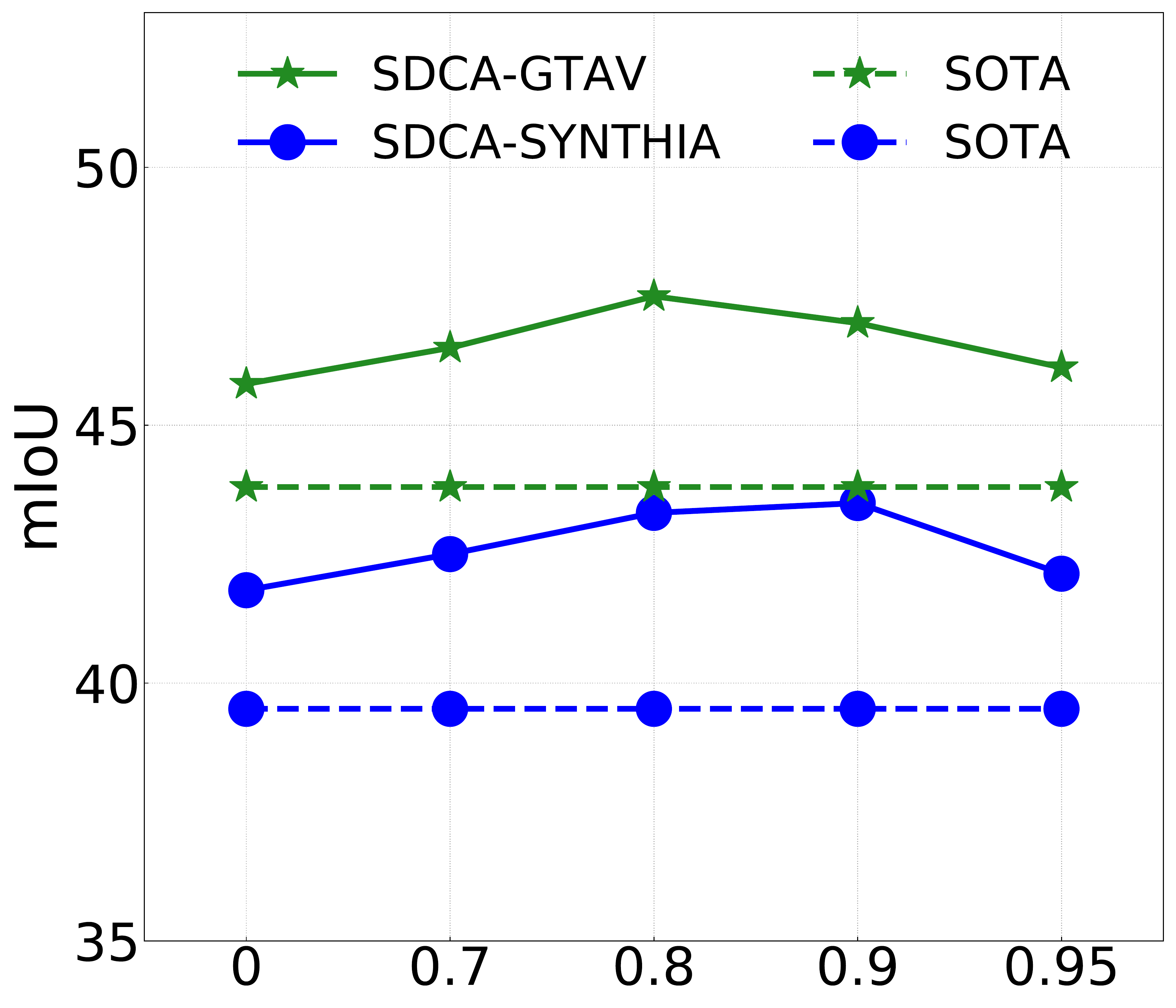}
        \label{Fig_parameter_threshold}
    }
    \subfigure[$\lambda_{lov}$]{
        \includegraphics[width=0.23\textwidth]{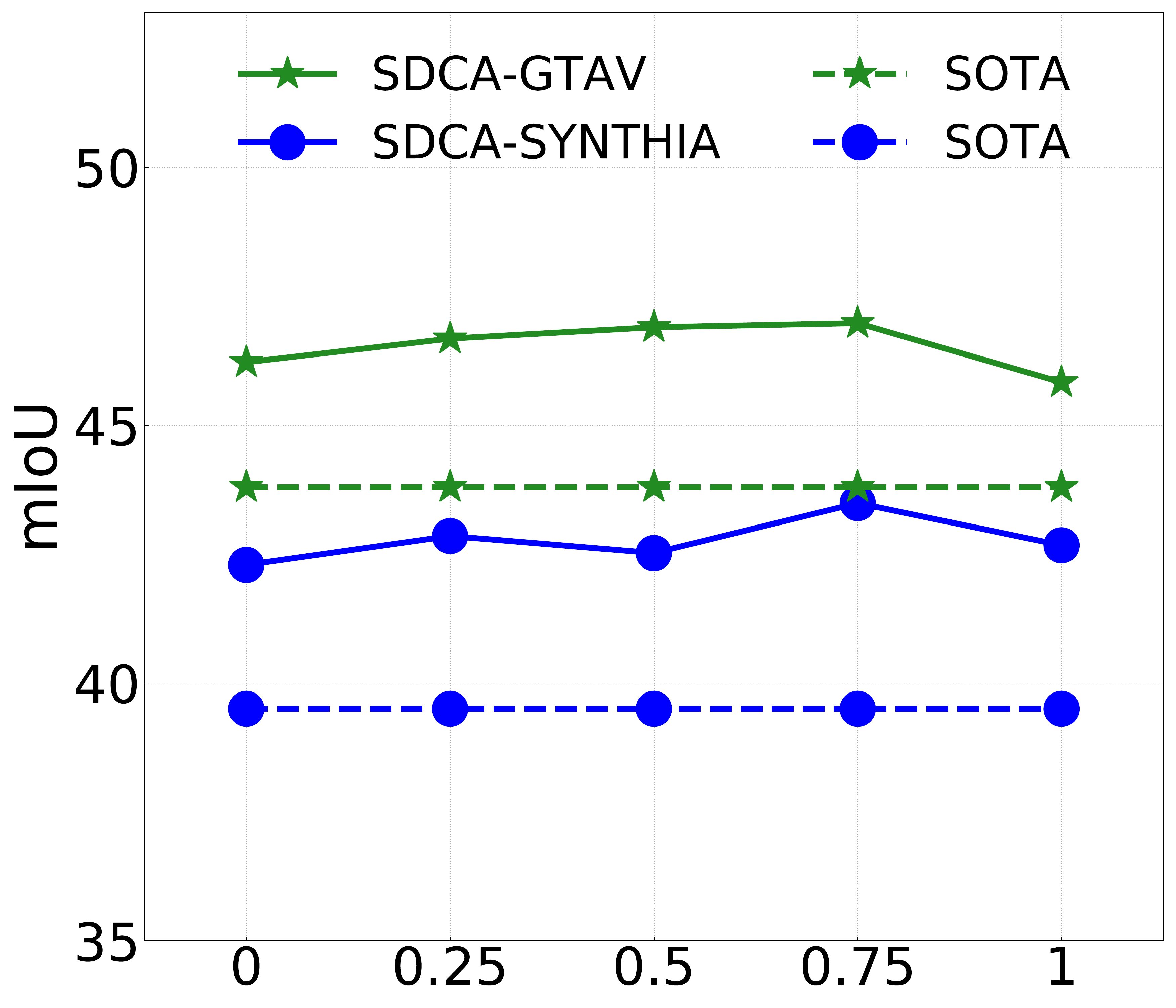}
        \label{Fig_parameter_lambda_lov}
    }
    \subfigure[$\lambda_{feat}$]{
        \includegraphics[width=0.23\textwidth]{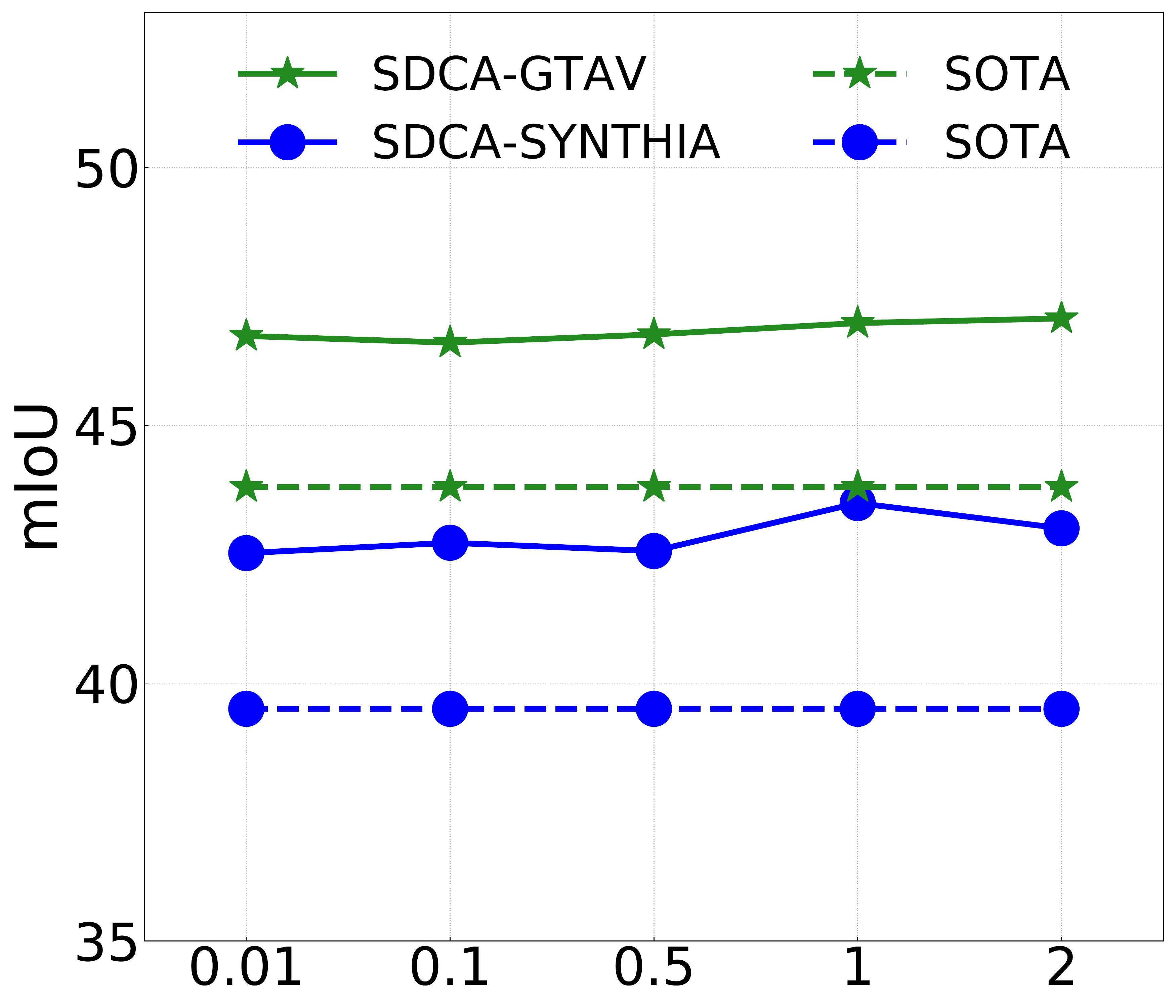}
        \label{Fig_parameter_lambda_feat}
    }
    \subfigure[$\lambda_{out}$]{
        \includegraphics[width=0.23\textwidth]{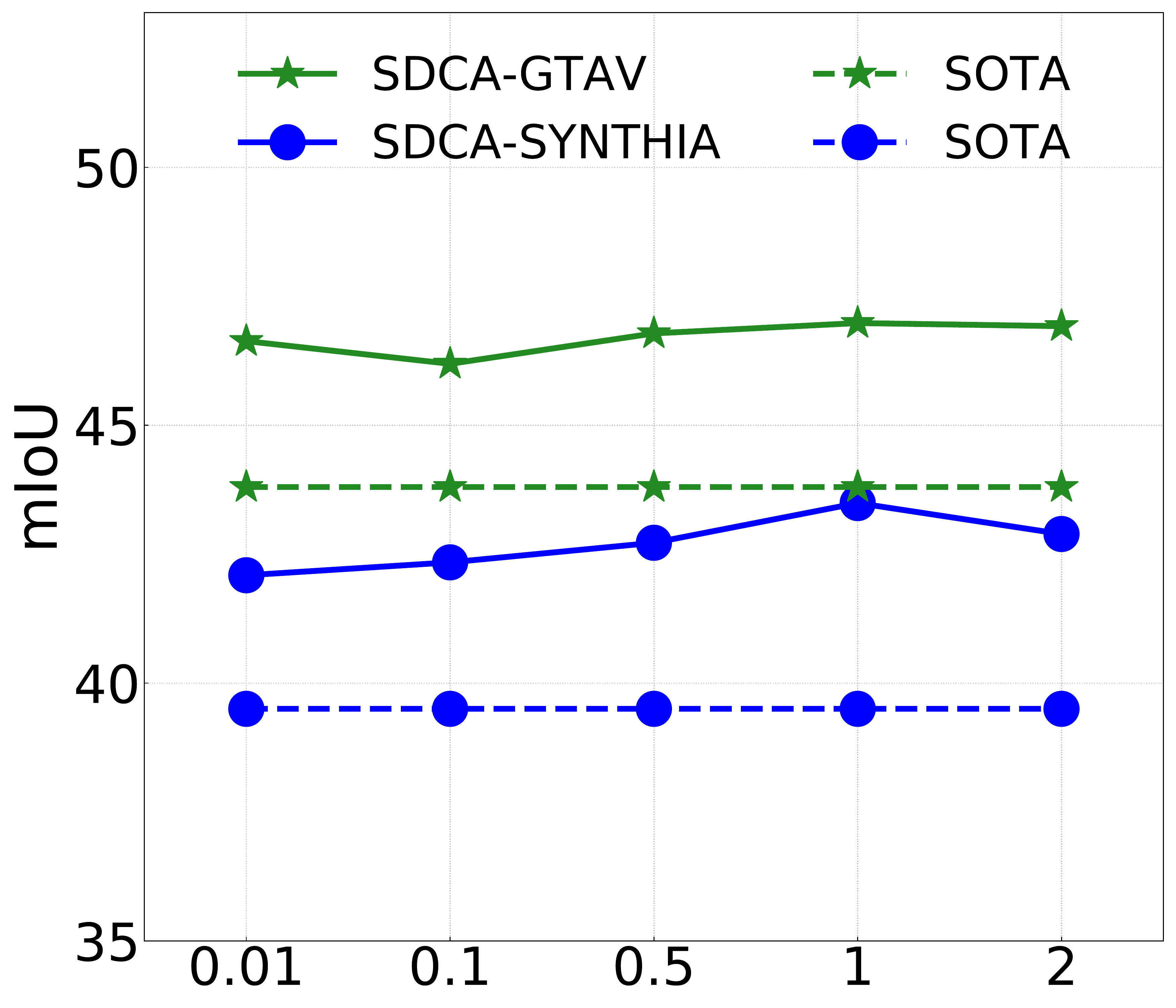}
        \label{Fig_parameter_lambda_pred}
    }
    \vspace{-2mm}
    \caption{Sensitivity analysis about hyper-parameters $\delta$, $\lambda_{lov}$, $\lambda_{feat}$ and $\lambda_{out}$. Previous state-of-the-art method (SOTA) and our SDCA are denoted as the dashed and solid curve respectively. The experiments are conducted on the scenarios of GTAV $\to$ Cityscapes (green) and SYNTHIA $\to$ Cityscapes (blue) with VGG-16 as the encoder.}
    \label{Fig_parameter_sensitivity}
    \vspace{-2mm}
\end{figure*}

\subsubsection{Pixel-wise Discrimination Distance}
To verify whether our semantic distribution-aware contrastive adaptation yields discriminative pixel representations, we design an experiment to take a closer look at what degree the pixel-wise representations are aligned. CLAN~\cite{luo2019taking} defines a Cluster Center Distance as the ratio of intra-category distance between the initial model and the aligned model while FADA~\cite{wang2020class} propose a new Class Center Distance to consider inter-category distance. To better evaluate the effectiveness of pixel-wise representation alignment, we introduce a new Pixel-wise Discrimination Distance (PDD) by taking intra- and inter-category affinities of pixel representations into account. Formally, the PDD for category $k$ is given by:
\begin{align}
    PDD(k) = \frac{1}{|\Lambda^k|} \sum_{x\in \Lambda^k} \frac{sim(x, \mu^k)}{\sum_{i=1,i\neq k}^K sim(x, \mu^{i})}
    \label{eq:pdd}
\end{align}%
where $sim(\cdot, \cdot)$ is the similarity metric and we adopt cosine similarity. $\Lambda^k$ denotes the pixel set that contains all the pixel representations belonging to the $k^{th}$ semantic class.

With PDD, we could investigate the relative magnitude of inter-category and intra-category pixel feature distances. Specifically, we calculate the PDD on the whole Cityscapes validate set and compare PDD values with other state-of-the-art category alignment methods: CLAN~\cite{luo2019taking} and FADA~\cite{wang2020class} for category-level adversarial training and SIM~\cite{wang2020differential} and CAG-UDA~\cite{zhang2019category} for category centroid based counterparts that do not tackle the distance between different category features. As shown in Fig.~\ref{Fig_ccd}, SDCA achieves a much higher PDD on most categories compared with other methods. Based on these quantitative results, together with the t-SNE analysis in Fig.~\ref{Fig_tsne}, it is clear that SDCA can achieve better pixel-wise category alignment and largely improve the pixel-wise accuracy of predictions.

\subsubsection{Parameter Sensitivity}
We conduct parameter sensitivity analysis to evaluate the sensitivity of SDCA on tasks GTAV $\to$ Cityscapes and SYNTHIA $\to$ Cityscapes. As shown in Fig.~\ref{Fig_parameter_sensitivity}, we select constant threshold $\delta \in \{0.0\,, 0.7\,, 0.8\,, 0.9\,, 0.95\}$ and balance weights $\lambda_{lov} \in \{0.0\,, 0.25\,, 0.5\,, 0.75\,, 1.0\}$, and $\lambda_{feat}\,,\lambda_{out} \in \{0.01\,, 0.1\,, 0.5\,, 1.0\,, 2.0\}$. For $\delta$, when $\delta$ is low, there are more pixels are involved in the process of alignment and when $\delta$ is high, there few pixels will be chosen. The mIoU of SDCA increases first and then decreases slightly. Similarly, as the values of $\lambda_{lov},\lambda_{feat},\lambda_{out}$ change in a vast range, the performance presents a slow bell-shaped curve and outperforms SOTA significantly on both tasks.

%% file: table/table_ablation_study.tex
\begin{table}
    \centering
    \caption{Ablation of the proposed loss functions for SYNTHIA $\to$ Cityscapes (ResNet-101).}\vspace{-2mm}
    \label{table:ablation_synthetic}
    \resizebox{0.48\textwidth}{!}{
    \begin{tabular}{ccccccc}
        \toprule[1.0pt]
        \multirow{2}{*}{Method} &\multicolumn{6}{c}{SYNTHIA $\rightarrow$ Cityscapes} \\
        \cmidrule(lr){2-7}
        & $\mathcal{L}_{lov}$ & $\mathcal{L}_{feat}$ & $\mathcal{L}_{seg}$ & $\mathcal{L}_{ssl}$  & mIoU & mIoU*  \\
        \midrule
        \multirow{3}{*}{Source Only} & {} & {} & {} & {} & 33.5 &  38.5\\ 
        & {\checkmark} & {} & {} & {} & 35.9 & 42.2 \\ 
        & {\checkmark} & {} & {} & {\checkmark} & 40.4 & 46.9 \\ 
        \midrule
        \midrule
        Ours (feat. only) & {\checkmark} & {\checkmark} & {} & {} & 43.5 & 50.8 \\  
        Ours (multi-level) & {\checkmark} & {\checkmark} & {\checkmark} & {} & 46.6 & 53.9 \\  
        Ours (full) & {\checkmark} & {\checkmark} & {\checkmark} & {\checkmark} & \bf 50.2 & \bf 56.8  \\  
         \bottomrule
    \end{tabular} 
    }
    \vspace{-2mm}
\end{table}

\begin{table}
    \centering
    \caption{Ablation of the proposed loss functions for GTAV $\to$ Cityscapes (ResNet-101).}\vspace{-2mm}
    \label{table:ablation_gta}
    \resizebox{0.48\textwidth}{!}{
    \begin{tabular}{ccccccc}
        \toprule[1.0pt]
        \multirow{2}{*}{Method} &\multicolumn{6}{c}{GTAV $\rightarrow$ Cityscapes} \\
        \cmidrule(lr){2-7}
        & $\mathcal{L}_{lov}$ & $\mathcal{L}_{feat}$ & $\mathcal{L}_{seg}$ & $\mathcal{L}_{ssl}$  & mIoU & mIoU$^t$  \\
        \midrule
        \multirow{3}{*}{Source Only} & {} & {} & {} & {} & 38.6 & 23.5 \\ 
        & {\checkmark} & {} & {} & {} & 40.7 & 26.5 \\ 
        & {\checkmark} & {} & {} & {\checkmark} & 45.3 & 30.6 \\ 
        \midrule
        \midrule
        Ours (feat. only) & {\checkmark} & {\checkmark} & {} & {} &  46.7 & 31.0 \\ 
        Ours (multi-level) & {\checkmark} & {\checkmark} & {\checkmark} & {} &  48.4 & 32.9 \\   
        Ours (full) & {\checkmark} & {\checkmark} & {\checkmark} & {\checkmark} &  \bf 52.9 & \bf 36.7 \\  
         \bottomrule
    \end{tabular} 
    }
    \vspace{-2mm}
\end{table}

%% file: 06-Conclusion.tex
\section{Conclusion}
\label{sec:conclusion}

In this paper, we propose a semantic distribution-aware contrastive adaptation (SDCA) method for semantic segmentation. The SDCA model successfully adapts the segmentation model to the target domain through pixel-wise alignment guided by semantic distributions. Specifically, we propose a particular form of contrastive loss at pixel level, which implicitly involves the joint learning of an infinite number of similar/dissimilar pixel pairs for each pixel-wise representation of both domains. Then, we derive an upper bound on this formulation and transfer the originally intractable loss function into practical implementation. This simple yet effective strategy works surprisingly well when coupled with self-supervised learning. The experimental results demonstrate the superiority of SDCA on various benchmarks.

%% file: main.bbl
\begin{thebibliography}{10}
\providecommand{\url}[1]{#1}
\csname url@samestyle\endcsname
\providecommand{\newblock}{\relax}
\providecommand{\bibinfo}[2]{#2}
\providecommand{\BIBentrySTDinterwordspacing}{\spaceskip=0pt\relax}
\providecommand{\BIBentryALTinterwordstretchfactor}{4}
\providecommand{\BIBentryALTinterwordspacing}{\spaceskip=\fontdimen2\font plus
\BIBentryALTinterwordstretchfactor\fontdimen3\font minus
  \fontdimen4\font\relax}
\providecommand{\BIBforeignlanguage}[2]{{%
\expandafter\ifx\csname l@#1\endcsname\relax
\typeout{** WARNING: IEEEtran.bst: No hyphenation pattern has been}%
\typeout{** loaded for the language `#1'. Using the pattern for}%
\typeout{** the default language instead.}%
\else
\language=\csname l@#1\endcsname
\fi
#2}}
\providecommand{\BIBdecl}{\relax}
\BIBdecl

\bibitem{geiger2012autonomous}
A.~Geiger, P.~Lenz, and R.~Urtasun, ``Are we ready for autonomous driving? the
  {KITTI} vision benchmark suite,'' in \emph{Proc. CVPR}, 2012, pp. 3354--3361.

\bibitem{zhou2020autonomous}
D.~Zhou, J.~Fang, X.~Song, L.~Liu, J.~Yin, Y.~Dai, H.~Li, and R.~Yang, ``Joint
  3d instance segmentation and object detection for autonomous driving,'' in
  \emph{Proc. CVPR}, 2020, pp. 1836--1846.

\bibitem{gupta2015scene}
S.~Gupta, P.~A. Arbel{\'{a}}ez, R.~B. Girshick, and J.~Malik, ``Indoor scene
  understanding with {RGB-D} images: Bottom-up segmentation, object detection
  and semantic segmentation,'' \emph{Int. J. Comput. Vis.}, vol. 112, no.~2,
  pp. 133--149, 2015.

\bibitem{li2009scene}
L.~Li, R.~Socher, and F.~Li, ``Towards total scene understanding:
  Classification, annotation and segmentation in an automatic framework,'' in
  \emph{Proc. CVPR}, 2009, pp. 2036--2043.

\bibitem{ronneberger2015UNet}
O.~Ronneberger, P.~Fischer, and T.~Brox, ``U-net: Convolutional networks for
  biomedical image segmentation,'' in \emph{Proc. MICCAI}, vol. 9351, 2015, pp.
  234--241.

\bibitem{mahapatra2013medical}
D.~Mahapatra, P.~J. Sch{\"{u}}ffler, J.~A.~W. Tielbeek, J.~Makanyanga,
  J.~Stoker, S.~A. Taylor, F.~M. Vos, and J.~M. Buhmann, ``Automatic detection
  and segmentation of crohn's disease tissues from abdominal {MRI},''
  \emph{{IEEE} Trans. Medical Imaging}, vol.~32, no.~12, pp. 2332--2347, 2013.

\bibitem{Cordts2016Cityscapes}
M.~Cordts, M.~Omran, S.~Ramos, T.~Rehfeld, M.~Enzweiler, R.~Benenson,
  U.~Franke, S.~Roth, and B.~Schiele, ``The cityscapes dataset for semantic
  urban scene understanding,'' in \emph{Proc. CVPR}, 2016, pp. 3213--3223.

\bibitem{stephan2016gtav}
S.~R. Richter, V.~Vineet, S.~Roth, and V.~Koltun, ``Playing for data: Ground
  truth from computer games,'' in \emph{Proc. ECCV}, 2016, pp. 102--118.

\bibitem{ros2016synthia}
G.~Ros, L.~Sellart, J.~Materzynska, D.~V{\'{a}}zquez, and A.~M. L{\'{o}}pez,
  ``The {SYNTHIA} dataset: {A} large collection of synthetic images for
  semantic segmentation of urban scenes,'' in \emph{Proc. CVPR}, 2016, pp.
  3234--3243.

\bibitem{dataset_shift_in_ML09}
J.~Quionero-Candela, M.~Sugiyama, A.~Schwaighofer, and N.~D. Lawrence,
  \emph{Dataset Shift in Machine Learning}.\hskip 1em plus 0.5em minus
  0.4em\relax The MIT Press, 2009.

\bibitem{ouyang2021progressive}
W.~Zhang, D.~Xu, J.~Zhang, and W.~Ouyang, ``Progressive modality cooperation
  for multi-modality domain adaptation,'' \emph{{IEEE} Trans. Image Process.},
  vol.~30, pp. 3293--3306, 2021.

\bibitem{jiebo2018deep}
F.~Long, T.~Yao, Q.~Dai, X.~Tian, J.~Luo, and T.~Mei, ``Deep domain adaptation
  hashing with adversarial learning,'' in \emph{Proc. SIGIR}, 2018, pp.
  725--734.

\bibitem{luo2021category}
Y.~{Luo}, P.~{Liu}, L.~{Zheng}, T.~{Guan}, J.~{Yu}, and Y.~{Yang},
  ``Category-level adversarial adaptation for semantic segmentation using
  purified features,'' \emph{{IEEE} Trans. Pattern Anal. Mach. Intell.}, pp.
  1--1, 2021.

\bibitem{pan2010survey}
S.~J. Pan and Q.~Yang, ``A survey on transfer learning,'' \emph{{IEEE} Trans.
  Knowl. Data Eng.}, vol.~22, no.~10, pp. 1345--1359, 2010.

\bibitem{liu2018corpus}
N.~Liu, Y.~Zong, B.~Zhang, L.~Liu, J.~Chen, G.~Zhao, and J.~Zhu, ``Unsupervised
  cross-corpus speech emotion recognition using domain-adaptive subspace
  learning,'' in \emph{Proc. ICASSP}, 2018, pp. 5144--5148.

\bibitem{liuli2019bow}
L.~Liu, J.~Chen, P.~W. Fieguth, G.~Zhao, R.~Chellappa, and
  M.~Pietik{\"{a}}inen, ``From bow to {CNN:} two decades of texture
  representation for texture classification,'' \emph{Int. J. Comput. Vis.},
  vol. 127, no.~1, pp. 74--109, 2019.

\bibitem{dundar2020stylization}
A.~{Dundar}, M.~Y. {Liu}, Z.~{Yu}, T.~C. {Wang}, J.~{Zedlewski}, and
  J.~{Kautz}, ``Domain stylization: A fast covariance matching framework
  towards domain adaptation,'' \emph{{IEEE} Trans. Pattern Anal. Mach.
  Intell.}, pp. 1--1, 2020.

\bibitem{yang2020fda}
Y.~Yang and S.~Soatto, ``{FDA:} fourier domain adaptation for semantic
  segmentation,'' in \emph{Proc. CVPR}, 2020, pp. 4084--4094.

\bibitem{li2019bidirectional}
Y.~Li, L.~Yuan, and N.~Vasconcelos, ``Bidirectional learning for domain
  adaptation of semantic segmentation,'' in \emph{Proc. CVPR}, 2019, pp.
  6936--6945.

\bibitem{hoffman2016fcns}
J.~Hoffman, D.~Wang, F.~Yu, and T.~Darrell, ``Fcns in the wild: Pixel-level
  adversarial and constraint-based adaptation,'' \emph{CoRR}, vol.
  abs/1612.02649, 2016.

\bibitem{luo2019significance}
Y.~Luo, P.~Liu, T.~Guan, J.~Yu, and Y.~Yang, ``Significance-aware information
  bottleneck for domain adaptive semantic segmentation,'' in \emph{Proc. ICCV},
  2019, pp. 6778--6787.

\bibitem{Hoffman_cycada2017}
J.~Hoffman, E.~Tzeng, T.~Park, J.-Y. Zhu, P.~Isola, K.~Saenko, A.~Efros, and
  T.~Darrell, ``{C}y{CADA}: Cycle-consistent adversarial domain adaptation,''
  in \emph{Proc. ICML}, 2018, pp. 1989--1998.

\bibitem{tsai2018learning}
Y.-H. Tsai, W.-C. Hung, S.~Schulter, K.~Sohn, M.-H. Yang, and M.~Chandraker,
  ``Learning to adapt structured output space for semantic segmentation,'' in
  \emph{Proc. CVPR}, 2018, pp. 7472--7481.

\bibitem{vu2019advent}
T.-H. Vu, H.~Jain, M.~Bucher, M.~Cord, and P.~P{\'e}rez, ``Advent: Adversarial
  entropy minimization for domain adaptation in semantic segmentation,'' in
  \emph{Proc. CVPR}, 2019, pp. 2517--2526.

\bibitem{pan2020unsupervised}
F.~Pan, I.~Shin, F.~Rameau, S.~Lee, and I.~S. Kweon, ``Unsupervised
  intra-domain adaptation for semantic segmentation through self-supervision,''
  in \emph{Proc. CVPR}, 2020, pp. 3764--3773.

\bibitem{goodfellow2014gan}
I.~J. Goodfellow, J.~Pouget{-}Abadie, M.~Mirza, B.~Xu, D.~Warde{-}Farley,
  S.~Ozair, A.~C. Courville, and Y.~Bengio, ``Generative adversarial nets,'' in
  \emph{Proc. NeurIPS}, 2014, pp. 2672--2680.

\bibitem{chen2017nomorediscrimination}
Y.~Chen, W.~Chen, Y.~Chen, B.~Tsai, Y.~F. Wang, and M.~Sun, ``No more
  discrimination: Cross city adaptation of road scene segmenters,'' in
  \emph{Proc. ICCV}, 2017, pp. 2011--2020.

\bibitem{wang2020class}
H.~Wang, T.~Shen, W.~Zhang, L.~Duan, and T.~Mei, ``Classes matter: {A}
  fine-grained adversarial approach to cross-domain semantic segmentation,'' in
  \emph{Proc. ECCV}, vol. 12359, 2020, pp. 642--659.

\bibitem{du2019ssf-dan}
L.~Du, J.~Tan, H.~Yang, J.~Feng, X.~Xue, Q.~Zheng, X.~Ye, and X.~Zhang,
  ``Ssf-dan: Separated semantic feature based domain adaptation network for
  semantic segmentation,'' in \emph{Proc. ICCV}, 2019, pp. 982--991.

\bibitem{kang2020pixel}
G.~Kang, Y.~Wei, Y.~Yang, Y.~Zhuang, and A.~G. Hauptmann, ``Pixel-level cycle
  association: {A} new perspective for domain adaptive semantic segmentation,''
  in \emph{Proc. NeurIPS}, 2020.

\bibitem{wang2020differential}
Z.~Wang, M.~Yu, Y.~Wei, R.~Feris, J.~Xiong, W.-m. Hwu, T.~S. Huang, and H.~Shi,
  ``Differential treatment for stuff and things: A simple unsupervised domain
  adaptation method for semantic segmentation,'' in \emph{Proc. CVPR}, 2020,
  pp. 12\,635--12\,644.

\bibitem{zhang2019category}
Q.~Zhang, J.~Zhang, W.~Liu, and D.~Tao, ``Category anchor-guided unsupervised
  domain adaptation for semantic segmentation,'' in \emph{Proc. NeurIPS}, 2019,
  pp. 433--443.

\bibitem{long2015fully}
J.~Long, E.~Shelhamer, and T.~Darrell, ``Fully convolutional networks for
  semantic segmentation,'' in \emph{Proc. CVPR}, 2015, pp. 3431--3440.

\bibitem{yu2016dilated}
F.~Yu and V.~Koltun, ``Multi-scale context aggregation by dilated
  convolutions,'' in \emph{Proc. ICLR}, 2016.

\bibitem{zhao2017pspnet}
H.~Zhao, J.~Shi, X.~Qi, X.~Wang, and J.~Jia, ``Pyramid scene parsing network,''
  in \emph{Proc. CVPR}, 2017, pp. 6230--6239.

\bibitem{chen2018deeplab}
L.~Chen, G.~Papandreou, I.~Kokkinos, K.~Murphy, and A.~L. Yuille, ``Deeplab:
  Semantic image segmentation with deep convolutional nets, atrous convolution,
  and fully connected crfs,'' \emph{{IEEE} Trans. Pattern Anal. Mach. Intell.},
  vol.~40, no.~4, pp. 834--848, 2018.

\bibitem{chen2018encoderdecoder}
L.~Chen, Y.~Zhu, G.~Papandreou, F.~Schroff, and H.~Adam, ``Encoder-decoder with
  atrous separable convolution for semantic image segmentation,'' in
  \emph{Proc. ECCV}, 2018, pp. 833--851.

\bibitem{liu2019autodeeplab}
C.~Liu, L.~Chen, F.~Schroff, H.~Adam, W.~Hua, A.~L. Yuille, and F.~Li,
  ``Auto-deeplab: Hierarchical neural architecture search for semantic image
  segmentation,'' in \emph{Proc. CVPR}, 2019, pp. 82--92.

\bibitem{tzeng2015simultaneous}
E.~Tzeng, J.~Hoffman, T.~Darrell, and K.~Saenko, ``Simultaneous deep transfer
  across domains and tasks,'' in \emph{Proc. ICCV}, 2015, pp. 4068--4076.

\bibitem{DRCN}
S.~Li, C.~H. Liu, Q.~Lin, , Q.~Wen, L.~Su, G.~Huang, and Z.~Ding, ``Deep
  residual correction network for partial domain adaptation,'' \emph{IEEE
  Trans. Pattern Anal. Mach. Intell.}, pp. 1--1, 2020.

\bibitem{long2015dan}
M.~Long, Y.~Cao, J.~Wang, and M.~I. Jordan, ``Learning transferable features
  with deep adaptation networks,'' in \emph{Proc. ICML}, vol.~37, 2015, pp.
  97--105.

\bibitem{liuli2019guest}
L.~Liu, M.~Pietik{\"{a}}inen, J.~Chen, G.~Zhao, X.~Wang, and R.~Chellappa,
  ``Guest editors' introduction to the special section on compact and efficient
  feature representation and learning in computer vision,'' \emph{{IEEE} Trans.
  Pattern Anal. Mach. Intell.}, vol.~41, no.~10, pp. 2287--2290, 2019.

\bibitem{ouyang2018collaborative}
W.~{Zhang}, D.~{Xu}, W.~{Ouyang}, and W.~{Li}, ``Self-paced collaborative and
  adversarial network for unsupervised domain adaptation,'' \emph{{IEEE} Trans.
  Pattern Anal. Mach. Intell.}, pp. 1--1, 2019.

\bibitem{long2016rtn}
M.~Long, H.~Zhu, J.~Wang, and M.~I. Jordan, ``Unsupervised domain adaptation
  with residual transfer networks,'' in \emph{Proc. NeurIPS}, 2016, pp.
  136--144.

\bibitem{long2017jan}
------, ``Deep transfer learning with joint adaptation networks,'' in
  \emph{Proc. ICML}, vol.~70, 2017, pp. 2208--2217.

\bibitem{huang2006correcting}
J.~Huang, A.~J. Smola, A.~Gretton, K.~M. Borgwardt, and B.~Sch{\"{o}}lkopf,
  ``Correcting sample selection bias by unlabeled data,'' in \emph{Proc.
  NeurIPS}, 2006, pp. 601--608.

\bibitem{gong2012geodesic}
B.~Gong, Y.~Shi, F.~Sha, and K.~Grauman, ``Geodesic flow kernel for
  unsupervised domain adaptation,'' in \emph{Proc. CVPR}, 2012, pp. 2066--2073.

\bibitem{DICD}
S.~Li, S.~Song, G.~Huang, Z.~Ding, and C.~Wu, ``Domain invariant and class
  discriminative feature learning for visual domain adaptation,'' \emph{{IEEE}
  Trans. Image Process.}, vol.~27, no.~9, pp. 4260--4273, 2018.

\bibitem{tzeng2014deep}
E.~Tzeng, J.~Hoffman, N.~Zhang, K.~Saenko, and T.~Darrell, ``Deep domain
  confusion: Maximizing for domain invariance,'' \emph{CoRR}, vol.
  abs/1412.3474, 2014.

\bibitem{gretton2012mmd}
A.~Gretton, B.~K. Sriperumbudur, D.~Sejdinovic, H.~Strathmann, S.~Balakrishnan,
  M.~Pontil, and K.~Fukumizu, ``Optimal kernel choice for large-scale
  two-sample tests,'' in \emph{Proc. NeurIPS}, 2012, pp. 1214--1222.

\bibitem{tzeng2017adversarial}
E.~Tzeng, J.~Hoffman, K.~Saenko, and T.~Darrell, ``Adversarial discriminative
  domain adaptation,'' in \emph{Proc. CVPR}, 2017, pp. 7167--7176.

\bibitem{ganin2015dann}
Y.~Ganin and V.~S. Lempitsky, ``Unsupervised domain adaptation by
  backpropagation,'' in \emph{Proc. ICML}, vol.~37, 2015, pp. 1180--1189.

\bibitem{long2018conditional}
M.~Long, Z.~Cao, J.~Wang, and M.~I. Jordan, ``Conditional adversarial domain
  adaptation,'' in \emph{Proc. NeurIPS}, 2018, pp. 1647--1657.

\bibitem{saito2018maximum}
K.~Saito, K.~Watanabe, Y.~Ushiku, and T.~Harada, ``Maximum classifier
  discrepancy for unsupervised domain adaptation,'' in \emph{Proc. CVPR}, 2018,
  pp. 3723--3732.

\bibitem{kang2019contrastive}
G.~Kang, L.~Jiang, Y.~Yang, and A.~G. Hauptmann, ``Contrastive adaptation
  network for unsupervised domain adaptation,'' in \emph{Proc. CVPR}, 2019, pp.
  4893--4902.

\bibitem{xie2018learning}
S.~Xie, Z.~Zheng, L.~Chen, and C.~Chen, ``Learning semantic representations for
  unsupervised domain adaptation,'' in \emph{Proc. ICML}, vol.~80, 2018, pp.
  5419--5428.

\bibitem{zou2018unsupervised}
Y.~Zou, Z.~Yu, B.~V. Kumar, and J.~Wang, ``Unsupervised domain adaptation for
  semantic segmentation via class-balanced self-training,'' in \emph{Proc.
  ECCV}, 2018, pp. 289--305.

\bibitem{Zhang_2017_ICCV}
Y.~Zhang, P.~David, and B.~Gong, ``Curriculum domain adaptation for semantic
  segmentation of urban scenes,'' in \emph{Proc. ICCV}, 2017, pp. 2039--2049.

\bibitem{zou2019confidence}
Y.~Zou, Z.~Yu, X.~Liu, B.~V. K.~V. Kumar, and J.~Wang, ``Confidence regularized
  self-training,'' in \emph{Proc. ICCV}, 2019, pp. 5982--5991.

\bibitem{zhang2020curriculum}
Y.~Zhang, P.~David, H.~Foroosh, and B.~Gong, ``A curriculum domain adaptation
  approach to the semantic segmentation of urban scenes,'' \emph{{IEEE} Trans.
  Pattern Anal. Mach. Intell.}, vol.~42, no.~8, pp. 1823--1841, 2020.

\bibitem{CycleGAN2017}
J.~Zhu, T.~Park, P.~Isola, and A.~A. Efros, ``Unpaired image-to-image
  translation using cycle-consistent adversarial networks,'' in \emph{Proc.
  ICCV}, 2017, pp. 2242--2251.

\bibitem{luo2019taking}
Y.~Luo, L.~Zheng, T.~Guan, J.~Yu, and Y.~Yang, ``Taking a closer look at domain
  shift: Category-level adversaries for semantics consistent domain
  adaptation,'' in \emph{Proc. CVPR}, 2019, pp. 2507--2516.

\bibitem{lian2019pycda}
Q.~Lian, L.~Duan, F.~Lv, and B.~Gong, ``Constructing self-motivated pyramid
  curriculums for cross-domain semantic segmentation: {A} non-adversarial
  approach,'' in \emph{Proc. ICCV}, 2019, pp. 6757--6766.

\bibitem{hadsell2006dimensionality}
R.~Hadsell, S.~Chopra, and Y.~LeCun, ``Dimensionality reduction by learning an
  invariant mapping,'' in \emph{Proc. CVPR}, 2006, pp. 1735--1742.

\bibitem{oord2018infoNCE}
A.~van~den Oord, Y.~Li, and O.~Vinyals, ``Representation learning with
  contrastive predictive coding,'' \emph{CoRR}, vol. abs/1807.03748, 2018.

\bibitem{he2020momentum}
K.~He, H.~Fan, Y.~Wu, S.~Xie, and R.~B. Girshick, ``Momentum contrast for
  unsupervised visual representation learning,'' in \emph{Proc. CVPR}, 2020,
  pp. 9726--9735.

\bibitem{chen2020contrastive}
T.~Chen, S.~Kornblith, M.~Norouzi, and G.~E. Hinton, ``A simple framework for
  contrastive learning of visual representations,'' in \emph{Proc. ICML}, vol.
  119, 2020, pp. 1597--1607.

\bibitem{cai2020jcl}
Q.~Cai, Y.~Wang, Y.~Pan, T.~Yao, and T.~Mei, ``Joint contrastive learning with
  infinite possibilities,'' in \emph{Proc. NeurIPS}, 2020.

\bibitem{chuang2020debiased}
C.~Chuang, J.~Robinson, Y.~Lin, A.~Torralba, and S.~Jegelka, ``Debiased
  contrastive learning,'' in \emph{Proc. NeurIPS}, 2020.

\bibitem{xie2020contrastive_dense}
Z.~Xie, Y.~Lin, Z.~Zhang, Y.~Cao, S.~Lin, and H.~Hu, ``Propagate yourself:
  Exploring pixel-level consistency for unsupervised visual representation
  learning,'' \emph{CoRR}, vol. abs/2011.10043, 2020.

\bibitem{wang2020contrastive_dense}
X.~Wang, R.~Zhang, C.~Shen, T.~Kong, and L.~Li, ``Dense contrastive learning
  for self-supervised visual pre-training,'' \emph{CoRR}, vol. abs/2011.09157,
  2020.

\bibitem{wang2021isda}
Y.~Wang, G.~Huang, S.~Song, X.~Pan, Y.~Xia, and C.~Wu, ``Regularizing deep
  networks with semantic data augmentation,'' \emph{{IEEE} Trans. Pattern Anal.
  Mach. Intell.}, pp. 1--1, 2021.

\bibitem{berman2018lovasz}
M.~Berman, A.~R. Triki, and M.~B. Blaschko, ``The lov{\'{a}}sz-softmax loss:
  {A} tractable surrogate for the optimization of the intersection-over-union
  measure in neural networks,'' in \emph{Proc. CVPR}, 2018, pp. 4413--4421.

\bibitem{david2010theory}
S.~Ben{-}David, J.~Blitzer, K.~Crammer, A.~Kulesza, F.~Pereira, and J.~W.
  Vaughan, ``A theory of learning from different domains,'' \emph{Mach.
  Learn.}, vol.~79, no. 1-2, pp. 151--175, 2010.

\bibitem{everingham2015IoU}
M.~Everingham, S.~M.~A. Eslami, L.~V. Gool, C.~K.~I. Williams, J.~M. Winn, and
  A.~Zisserman, ``The pascal visual object classes challenge: {A}
  retrospective,'' \emph{Int. J. Comput. Vis.}, vol. 111, no.~1, pp. 98--136,
  2015.

\bibitem{simonyan2015vgg}
K.~Simonyan and A.~Zisserman, ``Very deep convolutional networks for
  large-scale image recognition,'' in \emph{Proc. ICLR}, 2015.

\bibitem{he2016deep}
K.~He, X.~Zhang, S.~Ren, and J.~Sun, ``Deep residual learning for image
  recognition,'' in \emph{Proc. CVPR}, 2016, pp. 770--778.

\bibitem{deng2009imagenet}
J.~Deng, W.~Dong, R.~Socher, L.-J. Li, K.~Li, and L.~Fei-Fei, ``Imagenet: A
  large-scale hierarchical image database,'' in \emph{Proc. CVPR}, 2009, pp.
  248--255.

\bibitem{paszke2019pytorch}
A.~Paszke, S.~Gross, F.~Massa, A.~Lerer, J.~Bradbury, G.~Chanan, T.~Killeen,
  Z.~Lin, N.~Gimelshein, L.~Antiga, A.~Desmaison, A.~K{\"{o}}pf, E.~Yang,
  Z.~DeVito, M.~Raison, A.~Tejani, S.~Chilamkurthy, B.~Steiner, L.~Fang,
  J.~Bai, and S.~Chintala, ``Pytorch: An imperative style, high-performance
  deep learning library,'' in \emph{Proc. NeurIPS}, 2019, pp. 8024--8035.

\bibitem{maaten2008visualizing}
L.~van~der Maaten and G.~Hinton, ``Visualizing data using t-sne,'' \emph{J.
  Mach. Learn. Res.}, vol.~9, no.~86, pp. 2579--2605, 2008.

\end{thebibliography}
